\documentclass[11pt]{article}
\usepackage[preprint]{acl}
\usepackage{times}
\usepackage{latexsym}
\usepackage[T1]{fontenc}
\usepackage[utf8]{inputenc}
\usepackage{microtype}
\usepackage{inconsolata}
\usepackage{graphicx}
\usepackage{amsmath}
\usepackage{amssymb}
\usepackage{booktabs}
\usepackage{array}
\newcommand{\yes}{\checkmark}
\newcommand{\no}{$\times$}
\newcommand{\nd}{$\circ$}

\title{Many Solutions, One Answer:\\ Search Order Shapes the Default Solutions of Language Models}

\author{Qinhan Hou \and Jing Tang\\
  University of Helsinki, Helsinki, Finland \\
  \texttt{hou.qinhan@helsinki.fi \and jing.tang@helsinki.fi}
}

\begin{document}
\maketitle
\begin{abstract}
Language models homogenise their answers: when a problem has several correct ones, they tend to give the same one.
Whether the others are lost or merely not preferred decides whether this collapse can be undone.
We study six open-weight models on symbolic puzzles, ZebraLogic variants and three public planning tasks, all with known solution sets.
On every task a model completed, repeated samples concentrate on one default solution far beyond uniform sampling; all 15 model pairs share defaults on both puzzle tasks, and models usually produce rarely sampled solutions when clues single them out.
On held-out symbolic puzzles, with predictions fixed beforehand, a search that fills the entity with fewest candidates first matches more requests than name order in all six models (by 26 to 45 percentage points where the predictions differ), including two added after the explanation was developed; stated orders moved the outputs wherever we stated one.
On both puzzle tasks, randomising the presentation or search order broadens valid answers by 11 to 49 percentage points, more than asking for a random choice, and loses no validity beyond pre-set margins in 21 of 22 comparisons.
The order behind the default is task-specific: on calendars it is the earliest time, and shuffling does not broaden answers.
\end{abstract}

\section{Introduction}
\label{sec:introduction}

Many reasoning problems have more than one correct answer: a schedule can satisfy every constraint in several ways, and a puzzle can admit several completions.
Which one a language model (LM) returns matters because many people put the same question to a few models, so a model that always returns one valid answer narrows what all of them see.
Yet LMs tend to collapse this diversity: a model tends to return the same answer, and different models often return the same one~\citep{jiang_artificial_2026}, a homogenisation shaped by training and decoding choices~\citep{holtzman_curious_2020,kirk_understanding_2024,padmakumar_does_2024}.
A narrow output has two readings: the alternatives may be lost, or the model may prefer one answer while it can still produce the others, much as models encode factual answers they rarely generate~\citep{gekhman_inside-out_2025}.
Separating them, and finding what decides the preferred answer, shows whether the convergence can be undone while the answers stay valid.

\begin{figure}[t]
  \centering
  \includegraphics[width=\columnwidth]{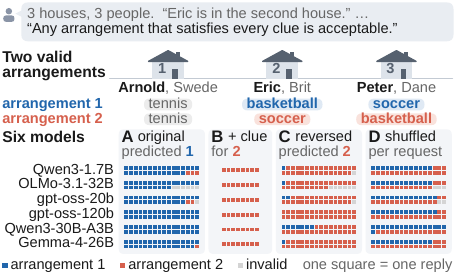}
  \caption{\textbf{Two valid arrangements, one default.}
  Prompt abridged (Appendix~\ref{app:tasks}).
  With the original prompt (A), all six models mostly return arrangement~1, which a house-by-house search, trying values in stated order, reaches first.
  A clue for arrangement~2 yields it (B), reversing value lists moves replies to it (C), and a random presentation per request splits them (D).}
  \label{fig:overview}
  \vspace{-3pt}%
\end{figure}

Open-ended generation makes this hard to test, because its set of correct answers is unknown.
We therefore study problems whose solution sets are known exactly (Sec.~\ref{sec:setting}): two \emph{controlled tasks}, a symbolic task that assigns five entities to five values with two, four or six solutions and multi-solution variants of ZebraLogic~\citep{lin_zebralogic_2025}, and multi-solution instances of three public planning benchmarks (Sec.~\ref{sec:benchmarks}).
Every valid output maps to one solution, so we separate lost from preferred alternatives by \emph{cued recoverability}: whether a model produces a solution when added facts leave only that solution.
A cue changes the problem, so it shows whether a solution can still be reached, not whether the model weighed it when answering freely.
We test six open-weight models from four developers and three training stages of Olmo~3 7B Think.
Where recent work on such problems compares whole solution distributions~\citep{wang_investigating_2026,daliberti_measuring_2026}, we predict the favoured solution before generation, test the prediction by changing orders, and cue each alternative.

Repeated samples concentrate on one \emph{default} solution far more than uniform sampling would, on every task a model completed (Figure~\ref{fig:overview}A); all 15 model pairs share defaults on the controlled tasks, and a model cued towards a rarely produced solution usually gives it (Figure~\ref{fig:overview}B; Sec.~\ref{sec:concentration}).
Which solution becomes the default is predicted by an ordered-search explanation (Sec.~\ref{sec:search}).
A search that takes the entities in name order and one that first fills the entity with the fewest candidate values can reach different assignments, and the defaults most often match the second, with values tried in their stated order.
We call this rule \emph{MRV-like}, after the minimum-remaining-values (MRV) heuristic of constraint solvers~\citep{haralick_increasing_1980}; it describes outputs, not internal computation.
On held-out problems, with predictions fixed before generation, outputs match it more often than name order in all six models, including two added after we developed the explanation, and a stated order moves the outputs wherever we stated one (Figure~\ref{fig:overview}C).

If the default is the first solution of an ordered search, varying the order across requests should broaden the answers (Figure~\ref{fig:overview}D; Sec.~\ref{sec:randomisation}).
On the controlled tasks, randomising the presentation or the search order does so, mostly without losing validity beyond margins set in advance, and more than asking for a random choice or raising the temperature; Verbalized Sampling broadens less on the symbolic task and for two of four models on ZebraLogic.
On three planning tasks from public benchmarks (Sec.~\ref{sec:benchmarks}), the order that sets the default is task-specific, and shuffling the presentation broadens answers on graphs and trips but not calendars.

\section{Related work}
\label{sec:related}

\paragraph{Output diversity and homogenisation.}
Sampling choices strongly affect the variety of generated text~\citep{holtzman_curious_2020}, preference tuning reduces output diversity relative to supervised fine-tuning~\citep{kirk_understanding_2024}, and writing with a model can homogenise content~\citep{padmakumar_does_2024}.
Different models converge on similar open-ended responses~\citep{jiang_artificial_2026,parikh_one-word_2026}, although the strength of this evidence is debated~\citep{schaeffer_how_2026}, and aligned models are less random than base models~\citep{west_base_2025} and less diverse than human writers~\citep{zhang_noveltybench_2025}.
With known answer spaces, \citet{hopkins_can_2023} score LM samples against target distributions, and on puzzles with three to eight solutions model solutions are less diverse than human ones and more alike across models~\citep{wang_investigating_2026}.
After reinforcement learning with verifiable rewards, base models match or overtake the tuned ones once many samples are allowed~\citep{yue_does_2025}, and the tuned models produce fewer of the solution modes a verifier distinguishes~\citep{daliberti_measuring_2026}; such collapse can be countered in training~\citep{puri_reaching_2026} or partly by prompting, as models asked for random outputs stay skewed~\citep{zhang_forcing_2024} and verbalising a distribution over responses recovers some diversity~\citep{zhang_verbalized_2026}, and it has been traced to how next-token probability is spread over valid and invalid continuations~\citep{banayeeanzade_sampling_2026}.
We count distinct final assignments, predict before generation the one each model favours, cue every alternative, and compare order randomisation with such baselines.

\paragraph{Order and presentation effects.}
LMs are sensitive to the order of few-shot examples~\citep{lu_fantastically_2022}, multiple-choice options~\citep{zheng_large_2024,pezeshkpour_large_2024}, premises~\citep{chen_premise_2024}, information in long contexts~\citep{liu_lost_2024}, the constraints of scheduling problems~\citep{sharma_schedbench_2026} and inputs more generally~\citep{guan_order_2025}, and the structural tokens of chat templates narrow open-ended outputs~\citep{yun_price_2025}.
These effects are usually averaged away as robustness failures; we predict which valid answer a change of order will favour and use randomised order to broaden the answers.

\begin{figure*}[t]
  \centering
  \includegraphics[width=\textwidth]{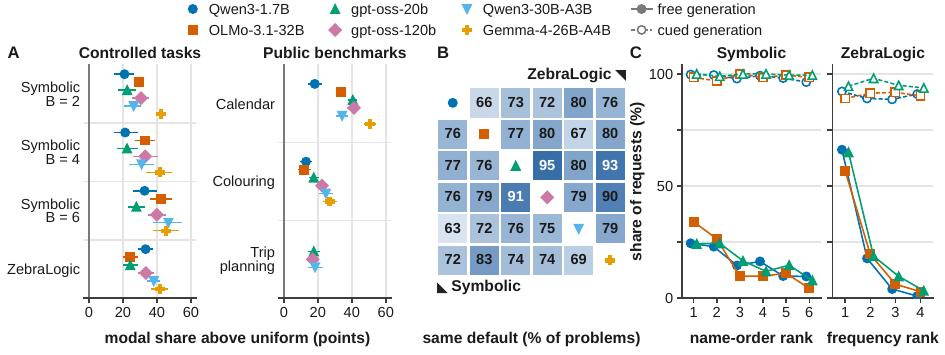}
  \caption{\textbf{Concentrated defaults and recoverable alternatives.}
  (A)~Modal share above uniform; intervals 95\% (benchmarks: 90\%).
  (B)~Problems (\%) where two models share a default (below diagonal Symbolic, above ZebraLogic; chance: 30\%, 39\%).
  (C)~Requests giving each solution freely (solid) or when cued (dashed), by rank; ZebraLogic is ranked by free frequency (original runs), so its solid curve must fall.}
  \label{fig:concentration}
\end{figure*}

\paragraph{Search and commitment in reasoning.}
Prompting and training methods make search explicit~\citep{yao_tree_2023,wang_self-consistency_2023,gandhi_stream_2024,lehnert_beyond_2024}.
Constraint-satisfaction solvers have long ordered variables by how few options remain~\citep{haralick_increasing_1980,russell_artificial_2020} and randomised such orders to vary their search~\citep{gomes_boosting_1998}; we use such procedures as behavioural predictors of the reported solution, not as claims about internal computation, which written reasoning need not reflect~\citep{turpin_language_2023} and which may be partly latent~\citep{hao_training_2026,geiping_scaling_2026}.
LMs can also struggle to revise conclusions~\citep{wilie_belief_2024}, become anchored to early answers in conversations~\citep{laban_llms_2025} and commit prematurely as agents~\citep{mehta_when_2026}; such findings fit both a model that can no longer use an alternative and one that merely prefers its earlier answer, which our cueing and history experiments separate.
NATURAL PLAN~\citep{zheng_natural_2024} scores a plan against one reference, and NPHardEval's colouring checker~\citep{fan_nphardeval_2024} accepts any proper colouring; we derive instances whose complete solution sets are known.

\section{Concentrated outputs, recoverable alternatives}
\label{sec:concentration}

\subsection{Setting and measurements}
\label{sec:setting}

In the \textbf{Symbolic} task, five entities $E_1,\dots,E_5$ each take a different value from $V_1,\dots,V_5$, and a list of exclusion facts (``$E_1$ is not in $V_2$'') leaves exactly $B$ valid assignments (Appendix Figure~\ref{fig:worked-example}; Appendix~\ref{app:tasks} gives a complete prompt).
We use 72 problems, 24 for each of $B\in\{2,4,6\}$: one from each of 24 structural classes at $B=4$ and 6, and relabelled instances of three classes at $B=2$; relabelled or reordered presentations are mapped back to a common solution set.
\textbf{ZebraLogic} variants pose the same questions in natural language (Figure~\ref{fig:overview} shows one): from 48 logic grid puzzles~\citep{lin_zebralogic_2025} of four sizes from $3\times3$ to $4\times4$, we removed clues until two, three or four solutions remained (24, 8 and 16 variants; Appendix Table~\ref{tab:zebra-provenance} gives the run and answer reader behind each result).

In \emph{free generation}, the prompt states that any solution satisfying the facts is acceptable, and we analyse the first 32 executed requests per problem and condition.
In \emph{cued generation}, added facts (on ZebraLogic, one added clue) leave only a target solution, and \emph{cued success} is the share of cued requests that produce it (Appendix~\ref{app:tasks}).
An output is \emph{valid} if it states one complete assignment that satisfies every constraint, and \emph{validity} is the share of all requests with a valid output, so invalid, ambiguous and truncated responses count against it.
Let $q(w)$ be the share of valid outputs that give solution $w$ among the $B$ solutions of a problem.
The \emph{modal share} $\max_w q(w)$ is compared with its expectation when the same number of valid outputs is drawn uniformly.
\emph{Breadth} is the effective number of solutions as a share of all $B$ solutions,
\begin{equation}
  \widetilde N_{\mathrm{eff}} = \frac{1}{B\sum_{w} q(w)^2},
  \label{eq:breadth}
\end{equation}
which is 100\% when valid outputs are spread evenly over all solutions and $100/B$\% when they all coincide~\citep{hill_diversity_1973}; with $B=4$, two equally frequent solutions give 50\%.
Breadth counts valid outputs only, so we report it with validity whenever we compare interventions, and from 32 requests it is biased downwards, so even uniform sampling falls short of 100\% (Appendix~\ref{app:zebra-residual}); differences are in percentage points, or points.

We developed the ordered-search explanation on three \emph{main models}, Qwen3-1.7B~\citep{yang_qwen3_2025} with thinking enabled, Olmo-3.1-32B-Instruct~\citep{olmo_olmo3_2025} (OLMo) and gpt-oss-20b~\citep{openai_gpt-oss_2025}; gpt-oss-120b tests a larger model, two \emph{new models}, Qwen3-30B-A3B with thinking enabled and Gemma-4-26B-A4B~\citep{gemma_team_gemma4_2026}, test the explanation on models it was not developed on, and the SFT, DPO and RLVR checkpoints of Olmo~3 7B Think test training stages.
All use their recommended decoding except OLMo, sampled at temperature 1.0 and top-$p$ 1.0, the setting we had fixed for models without a thinking mode (Sec.~\ref{sec:robustness}).
Intervals are 95\% bootstrap intervals over clusters of related problems~\citep{efron_introduction_1993}, structural classes or source puzzles, and tests are Holm-corrected within each model's family~\citep{holm_simple_1979}.
A validity change passes when the lower limit of its 90\% interval lies above $-3$ points on the Symbolic task and $-5$ points elsewhere; Appendix~\ref{app:stats} gives the two-sided equivalence rules~\citep{schuirmann_comparison_1987} that the earlier studies had specified.
We call a test \emph{pre-specified} when its analysis plan and the model predictions were committed to version control before the confirmatory run; Appendix Tables~\ref{tab:prespecified} and~\ref{tab:prespecified-b} list these tests, and the rest is descriptive.

\subsection{Concentration within and across models}
\label{sec:concentration-results}

We call the most frequent valid solution in 32 requests the model's \emph{default}.
All six models concentrate on it far more than uniform sampling would, on every task they completed (Figure~\ref{fig:concentration}A).
On the Symbolic task, the default's share of valid outputs exceeds the uniform expectation (about 57\% at $B=2$ and 26\% at $B=6$) by 21 to 42 points at $B=2$ and at $B=4$ and by 28 to 47 points at $B=6$, with every interval above zero.
Over all 48 ZebraLogic puzzles the excess is 24 to 42 points, and on the three planning tasks 12 to 51 points; each pre-specified test passes in every model that completed the task (Sec.~\ref{sec:benchmarks}).

Different models also share the default (Figure~\ref{fig:concentration}B; ties counted fractionally).
Across the 15 pairs, two models' defaults coincide on 63 to 91\% of Symbolic problems and 66 to 95\% of ZebraLogic puzzles, where about 30\% and 39\% would be expected if one of them chose uniformly among the valid solutions.
The agreement is reliably above this chance level in every pair (Appendix Table~\ref{tab:shared-defaults}; six of the comparisons are post hoc).
In the main pairs, relabelling the values moves both defaults to the same new solution on 64 to 75\% of problems and reversing the order of the entities' facts on only 13 to 17\% (chance 22 to 27\% and 5 to 6\%; Appendix~\ref{app:figure-notes}), as Sec.~\ref{sec:search} explains.

\subsection{Low frequency does not imply low recoverability}
\label{sec:recoverability}

Figure~\ref{fig:concentration}C compares how often a main model gives each solution freely with how often it gives that solution when cued towards it, both as shares of all requests (Appendix Figure~\ref{fig:appendix-rank} shows every $B$).
Every solution, the default included, is cued by the smallest set of added assignments that leaves only it (one for 197 of the 288 Symbolic targets, two or three for the rest).
On the Symbolic task, mean free-generation frequency is only 4 to 10\% at the last rank for $B=6$, whereas mean cued success stays at 95 to 100\% at every rank and $B$.
A pre-specified test places the slope of cued success over rank within $\pm5$ points per step for OLMo and gpt-oss-20b, and in each of the two new models all 2,304 cued requests succeeded, on default and non-default targets alike.
On ZebraLogic, non-default solutions are cued about as successfully as the default in all six models (differences from $-3.6$ to $+1.9$ points, within the pre-specified $\pm10$-point margin).
A rarely chosen solution is therefore not out of reach, even when the conversation pushes the model towards another: keeping an earlier answer in the conversation raises OLMo's free share of that answer by 15 to 33 points, yet cued success for the other solutions changes by less than 1 point (Appendix~\ref{app:history}).
A cue adds facts and so eases the problem, most at $B=6$, where half of the targets need more than one assignment. But alternatives are also reached without a cue: the order changes of Secs.~\ref{sec:search} and~\ref{sec:randomisation} add no facts, and asked only for a different solution, the four earlier models covered 56 to 75\% of the solution sets by the second turn, though the new answer continued their ordered search in only three of eight cases (Appendix~\ref{app:it}).
On the public benchmarks, recoverability varies by benchmark and model (Sec.~\ref{sec:benchmarks}).

\section{Ordered search predicts the default solution}
\label{sec:search}

If rarely produced solutions remain recoverable, something other than ability decides which one becomes the default; we test whether models behave as if they searched in a fixed order and reported the first complete solution.

\begin{figure*}[t]
  \centering
  \includegraphics[width=\textwidth]{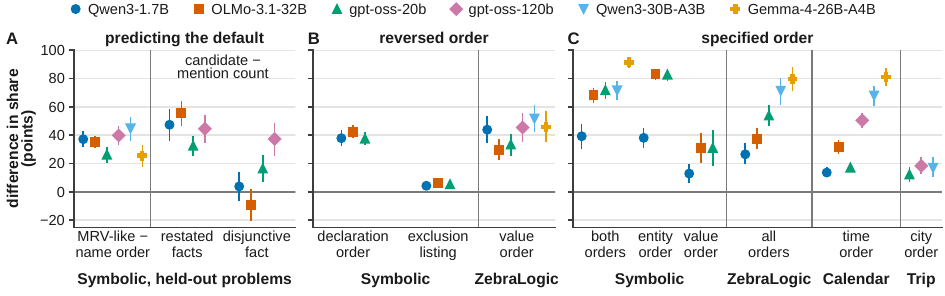}
  \caption{\textbf{Testing the ordered-search explanation.}
  (A)~MRV-like minus name-order share on held-out problems; candidate-count minus mention-count share on problems with restated facts or a disjunctive fact.
  (B,~C)~Change in the predicted solution's share when an order is reversed (B) or specified (C).
  Shares of all requests; 95\% intervals (benchmarks: 90\%).}
  \label{fig:search}
  \vspace{10.9pt}%
\end{figure*}

\subsection{Predicting the first solution}
\label{sec:predict}

A depth-first search fills one entity at a time, tries its remaining values in their stated order, backs up when no value fits and stops at the first complete solution; such searches differ mainly in the order in which they visit entities.
\emph{Name order} visits $E_1,\dots,E_5$ in turn, whereas \emph{fewest candidates first} visits entities in order of how many values the facts leave them, counted once before the search starts: a static form of MRV, whose dynamic form would recount after each assignment (Appendix Figure~\ref{fig:worked-example} shows how the two diverge).
Since each exclusion fact removes one of an entity's five values, the entity with the fewest candidates is also named most often; a design below decouples the two.

The hypothesis came from earlier free-generation data on the 72 main problems, where outputs matched MRV more often than name order by 30 to 44 points.
We used these data to select each model's primary procedure among eight candidates (static MRV in every case) and to fit the procedure mixture described below (Appendix~\ref{app:procedures}).
We then drew 48 new problem instances (20 from structural classes not used before), showed each in four presentations, and fixed every procedure's predicted solution before generation (Appendix~\ref{app:prospective}); the two new models, added later, selected static MRV on their own earlier answers and committed their predictions before any held-out response was generated.
On the 149 to 154 of 192 presentations where a model's primary procedure and name order predict different solutions, the share of all requests matching the MRV-like prediction exceeds the share matching name order in every model: by 37, 36 and 26 points for the main models (42 to 48\% of requests against 11 to 16\%), 40 for gpt-oss-120b, and 45 and 26 for Qwen3-30B-A3B and Gemma (Figure~\ref{fig:search}A; every pre-specified test passes).
The MRV-like prediction is also the most frequent valid answer on 63 to 69\% of the 192 held-out presentations in every model, against 19 to 35\% for name order and about 21\% for a uniform guess (post hoc; Appendix~\ref{app:prospective}).
On the 20 problems from new structural classes it is 25 to 38 points in each earlier model (post hoc, intervals above zero; Appendix~\ref{app:prospective}).
A further design on 42 new structural classes decouples candidate count from mention count (Figure~\ref{fig:search}A; Appendix~\ref{app:procedures}).
When restated facts add mentions of an entity without changing its candidates, outputs follow the candidate-count prediction rather than the mention-count one in all four earlier models (by 33 to 56 points); when a disjunctive fact (``$E_4$ is $V_1$ or $V_5$'') gives one entity both the fewest candidates and the fewest mentions, only the two gpt-oss models favour candidate count (by 17 and 37 points); OLMo leans towards mention count and Qwen3-1.7B shows no difference, both with intervals that include zero.
The rule thus often picks out the default on new problems, but whether candidates or mentions set the order, and finer distinctions between procedures (static or dynamic MRV, tie-breaking), were not consistent across models, so we call the default \emph{MRV-like} and read the procedures as predictors of behaviour, not as the computation inside the model.
Nor do they explain every answer: in a mixture of the eight procedures and uniform choice fitted to the free-generation answers, the procedures together carry 0.53, 0.72 and 0.52 of the weight for the main models and 0.57 and 0.80 for the two new ones, and the rest behaves like uniform choice (Appendix Table~\ref{tab:mixture}); the written reasoning fits a search without establishing one (Appendix~\ref{app:process}).

\subsection{Declaration order steers solution choice}
\label{sec:declaration}

An ordered search tries values in their stated order, so stating them in a different order should move the default.
To separate this order from familiar label orders such as $V_1<V_2$, we replaced the value labels with arbitrary words and declared them in one order or its reverse, on the 48 problems with $B=4$ or 6, listing each entity's exclusions in the declared order.
Reversing the two orders together raised the share on the solution the primary procedure reaches first under the new order by 42, 49 and 43 points, averaged over both directions, and a factorial design that crossed them (Figure~\ref{fig:search}B; Appendix~\ref{app:figure-notes}) shows a large effect of declaration order (38 to 42 points) and a small but reliable one of the exclusion-listing order, the order of the excluded values within each entity's facts (4 to 7 points, all intervals above zero).
This asymmetry is what the explanation predicts: a search must check every constraint wherever it is listed, but it tries values in the order in which they are declared.
On ZebraLogic, reversing the declared order of each attribute's values likewise raised the share on the newly predicted solution, by 30 to 52 points in all six models (Figure~\ref{fig:search}B; pre-specified).

\subsection{Specifying the search order}
\label{sec:specify}

The most direct test is to tell the model which order to use: to consider the entities in a given order, to try each entity's values in a given order, or both, and to keep the first choice that leads to a complete solution (Appendix~\ref{app:tasks}), with orders whose first solution differs from both the MRV-like and the name-order prediction.
Relative to the unmodified prompt, every instruction moved the Symbolic outputs towards the specified solution (Figure~\ref{fig:search}C): by 39 to 91 points when both orders were specified (main and new models), and in the main models by 38 to 83 points for the entity order alone and 13 to 31 points for the value order alone.

On ZebraLogic, where the instruction fixes the house order, an attribute order within each house and a value order for each attribute, it moved the outputs towards the specified solution by 27 to 80 points in the main and new models (all intervals above zero).
Among the four earlier models, a house-by-house search predicts a shared default in 155 of the 215 puzzle--pair comparisons in which two models share one (72\%, against 40\% by chance), but house-, attribute- and fewest-candidates-first searches return the same solution on 41 of the 48 puzzles, so the defaults cannot distinguish these procedures.
On both testbeds a stated order thus shifts which solution the models return, and it does so on the calendar and trip-planning benchmarks too (Sec.~\ref{sec:benchmarks}).

\subsection{Across scale, training stages and protocols}
\label{sec:robustness}

The pattern persists across the sizes we tested (Qwen3 1.7B and 30B, gpt-oss 20b and 120b), with size confounded with design and training (Appendix~\ref{app:stages}).
In Olmo~3 7B Think, the SFT checkpoint already concentrates and is recovered by cues; the stronger concentration after DPO comes with a change of system prompt, with the prompt fixed the checkpoint effect is not resolved, and RLVR adds no resolved change (Appendix~\ref{app:stages}).
Reduced reasoning keeps concentration and the MRV-like contrast, no written reasoning cuts validity by 57 to 86 points (Appendix~\ref{app:answer-only}), and OLMo's results hold at its recommended decoding (Appendix~\ref{app:models}).

\section{Broadening answers by varying order}
\label{sec:randomisation}

If a model tends to return the first solution it finds, varying the order across requests should broaden its answers. We compare changing the order with asking for randomness, on breadth and validity.

\begin{figure*}[t]
  \centering
  \includegraphics[width=\textwidth]{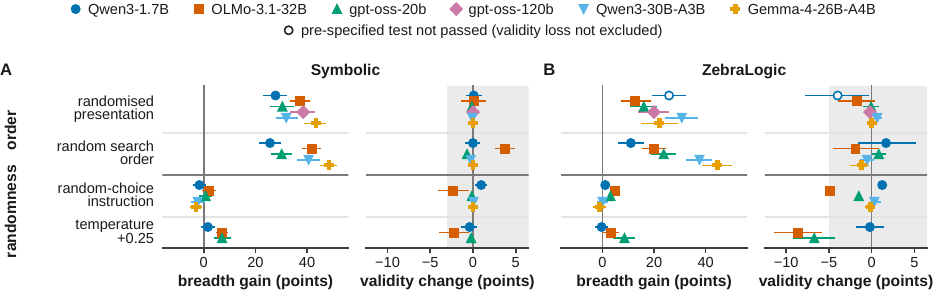}
  \caption{\textbf{Breadth and validity under each intervention.}
  Change from the unmodified prompt, in percentage points, on Symbolic (A) and ZebraLogic (B).
  Breadth counts valid answers, validity all requests; intervals 95\% (breadth) and 90\% (validity); shading: validity changes within the one-sided margin.
  The randomness rows are descriptive; ZebraLogic higher temperature from original runs.}
  \label{fig:interventions}
  \vspace{5.5pt}%
\end{figure*}

\subsection{Prompt and sampling conditions}
\label{sec:interventions}

\emph{Randomised presentation} (prompts in Appendix~\ref{app:tasks}) shows each request with the values randomly relabelled and the entities' facts in a random order (on ZebraLogic: random orders of each attribute's values, the attribute lines and the clues), and maps the answers back before scoring; it changes several things at once, and a variant also renames the entities.
A \emph{random search order} adds the instruction of Sec.~\ref{sec:specify}, with entity and value orders that we draw uniformly at random for each request and write into the prompt, so the model is not asked to invent an order.
To request randomness, a \emph{random-choice instruction} asks the model to choose one of the valid solutions uniformly at random, a \emph{higher temperature} adds 0.25 to the sampling temperature, and, on the Symbolic task, \emph{enumerate, then choose} asks for all solutions and a random choice among them, which changes the solving procedure.
We also ran two stronger baselines: \emph{Verbalized Sampling}~\citep{zhang_verbalized_2026}, in which the model states five candidate answers with probabilities and one is drawn accordingly, and the highest of two raised temperatures that kept validity within the margin.

For randomised presentation and the random search order we fixed a test in advance: it passes when breadth (Eq.~\ref{eq:breadth}) increases and the validity change passes the rule of Sec.~\ref{sec:setting}; the other comparisons are descriptive unless stated otherwise.
Of the 22 such tests on the two tasks, 21 meet this one-sided criterion and 19 the criteria originally specified for each study, which were two-sided for the earlier ones (Appendix~\ref{app:stats}).
Before the runs, we also forecast each intervention's breadth from the procedure mixture of Sec.~\ref{sec:predict} (Appendix~\ref{app:procedures}).

\subsection{Breadth and validity on the Symbolic task}
\label{sec:symbolic-interventions}

Both order interventions broaden the answers substantially (Figure~\ref{fig:interventions}A): randomised presentation raises breadth by 28 to 43 points in all six models and passes the test in each.
Renaming the entities as well gives similar gains in the main models (29 to 41 points; Appendix Figure~\ref{fig:appendix-interventions}).
A random search order raises breadth by 26 to 49 points in the five models we ran it with and passes in each.

Asking for randomness does much less.
The random-choice instruction changes breadth by $-3$ to $+2$ points in the five models it was run with (for Gemma a decrease, with an interval below zero), in line with the skewed outputs of models asked for random choices~\citep{zhang_forcing_2024}, and a pre-specified test confirms that randomised presentation broadens the answers more in each, by 29 to 46 points.
A higher temperature adds 2 to 7 points in the main models (a larger increase left OLMo and gpt-oss-20b with almost no valid answers; Appendix~\ref{app:temperature}).
Enumerating before choosing raises breadth for OLMo and gpt-oss-20b (by 24 and 22 points), lowers it for Qwen3-1.7B (by 10) and costs OLMo 20.5 points of validity (Appendix Figure~\ref{fig:appendix-interventions}).
Randomised presentation also broadens more than Verbalized Sampling (by 23 to 24 points) and the validity-preserving temperature (by 27 to 34 points) in each main model (Appendix~\ref{app:baselines}).
This contrast fits the ordered-search explanation: randomising the order changes the prompt itself, whereas asking for a random choice leaves the order, and hence the first solution, unchanged.

\subsection{Transfer to ZebraLogic}
\label{sec:zebra-interventions}

Randomised presentation raises breadth by 13 to 31 points in all six models (Figure~\ref{fig:interventions}B) and passes the pre-specified test in all but Qwen3-1.7B, whose validity falls by 4.0 points ([$-7.7$, $-0.3$]); it broadens more than the random-choice instruction in each of the five models tested, by 8 to 31 points.
A random search order raises breadth by 11 to 44 points in the five models we ran it with and passes in each.
Verbalized Sampling broadens less than randomised presentation for Qwen3-1.7B and OLMo but not resolvably so for the gpt-oss models, and its answers are misread more often (Appendix~\ref{app:baselines}).
More concentration remains after randomised presentation than the procedure mixture forecasts, so the defaults depend on more than the orders it permutes: house numbering, clue wording and value names stay unchanged, and the source puzzle's solution does not explain the rest (Appendices~\ref{app:zebra-residual} and~\ref{app:zebra-source}).

\section{Public planning benchmarks}
\label{sec:benchmarks}

To test whether the findings carry beyond our constructed tasks, we derived instances with two to six valid answers from three planning tasks of two public benchmarks (Appendix~\ref{app:benchmarks}): NATURAL PLAN calendar scheduling (72 instances, without the request for the earliest slot) and trip planning (48, with one event constraint removed and the scorer's rules stated in the prompt, without which no model produced a valid itinerary)~\citep{zheng_natural_2024}, and NPHardEval graph colouring (48 graphs with pre-coloured vertices, run without stated orders or the random-choice instruction)~\citep{fan_nphardeval_2024}.
For each we fixed in advance which order should set the default: calendar order for meeting times, fewest available colours first (static MRV) against highest degree first for colourings, and fewest feasible positions first for the cities of an itinerary (Appendix Table~\ref{tab:benchmark-estimates} gives every test).

Every admitted model concentrates on a default on all three tasks (Figure~\ref{fig:concentration}A), and listing candidate times or cities in a stated order moves the outputs towards the solution that order reaches first in every model run (Figure~\ref{fig:search}C).
Which order sets the default, however, depends on the task.
On calendars, the earliest feasible time predicts the default in five of six models; Qwen3-1.7B concentrates as well, but away from it (its share on the earliest time is 23 points below chance).
In graph colouring, fewest colours first beats highest degree first for the two gpt-oss models (by 17 points each) and the two are not distinguished for the other four; Gemma leans towards degree-first, unresolved after correction (adjusted $p=0.08$).
In trip planning, fewest feasible positions first does not predict the default in any of the three models admitted to it (Appendix~\ref{app:benchmarks}), but the benchmark's own prompts write each flight pair with the earlier city of the reference itinerary first, and keeping that order rather than a random one raises the reference itinerary's share by 10 to 12 points in all three.
Cued success on non-default solutions stays within 10 points of the default's in three of six, five of six and two of three models; a miss means only that the margin was not shown (Qwen3-30B-A3B on trips: $-8$ points).

Shuffling the presentation broadens the answers on graphs and trips but not on calendars.
Shuffling a graph passes the joint breadth-and-validity test in four of six models (for Qwen3-1.7B and OLMo a validity loss beyond 5 points is not excluded), and shuffling a trip's constraints and flights passes in all three admitted models.
On calendars, shuffling participants, busy intervals and days does not broaden the answers in any model.
A random order of candidate times does, by 14 to 53 points in all six models, but with too large a validity loss in four of them; a random city order passes in two of three models.
A random order also broadens more than the random-choice instruction in five of six models on calendars and two of three on trips, although for gpt-oss-20b on calendars the instruction broadens more (by 10 points).
This fits the idea that shuffling the text helps only where the default's order comes from the text, which we did not test.

\section{Discussion and conclusion}
\label{sec:conclusion}

Across six models and five tasks, the models favoured one answer when asked freely, often the same one across models, yet usually produced other valid answers when clues identified them, and an ordered-search rule fixed in advance predicted the favoured answer on new symbolic problems, also in models added after the rule was developed.
The rule predicts which answer a model reports, not how the model computes it, and the order that matters has to be identified for each task: on calendars, for instance, it is the earliest time.
For the debate on homogenisation, this has three implications.
Measures of output diversity should be read alongside tests of what a model can still produce, because free answers converge while the alternatives stay within reach.
Some homogeneity across models may come from shared problem-solving conventions applied to the same structure, such as trying values in the order stated or, on symbolic puzzles, filling the most constrained entity first, complementing token-level explanations~\citep{banayeeanzade_sampling_2026}.
And the convergence can be partly undone without access to model weights, by changing the order of presentation or search.

In practice, defaults are to be expected whenever a prompt lists options, values or constraints.
Asking again with the presentation shuffled is a cheap check: it broadened the answers on the controlled tasks, graphs and trips, though not on calendars, and it makes answers less reproducible.
Shuffling requires mapping answers back to the original labels; a random search order does not, and it broadened calendar answers too, but often at a cost in validity, and it lengthened ZebraLogic outputs by 14 to 72\%.
Open-ended tasks with many valid answers, such as code, are a natural next step.

\section*{Limitations}

Our conclusions are limited first by the tasks.
Each problem has two to six solutions that can be enumerated, which is what lets us check every answer and cue every alternative, but we have not tested whether the same explanation holds for open-ended generation, where the valid answers are many and unknown.

The ordered-search explanation is also partial.
It predicts which answer a model reports and says nothing about how the model computes it, and the search procedures carry only 52 to 80\% of the fitted answer distribution (Appendix~\ref{app:procedures}).
In the same way, a successful cue shows only that a solution can be reached once added facts single it out.
Randomising the order has a cost of its own, because the answers become broader but also harder to reproduce.

The set of models is small.
We tested six open-weight models from four developers and one Olmo~3 7B training series; we could not evaluate base checkpoints, the public benchmarks admitted only three models to trip planning, and closed models may behave differently.
Not every model ran every condition (Appendix Table~\ref{tab:studies} shows which models ran which tests), and each model ran under its own chat template and default system message, so our claims rest on comparisons between conditions within a model, not on how large an effect is in one model compared with another.
Four of the six are mixture-of-experts models, and all of them wrote out their reasoning (in a thinking or reasoning mode, or for OLMo in the answer itself), so what we can say about behaviour without written reasoning rests on the reduced-reasoning and answer-only checks (Appendix~\ref{app:answer-only}).

Some of the statistical evidence is weaker than a fully pre-specified analysis would give.
Several earlier validity decisions use a one-sided rule that we adopted after those studies were run; it changes four decisions, in each of which validity may have risen beyond the upper margin of the original two-sided test (Appendix~\ref{app:stats}).
The shared-default comparisons among the four earlier models are post hoc, and 16 test decisions changed, all towards support, after a correction made once results were available: the four just named, and twelve after corrections of analysis code that had returned an estimate of exactly zero or tested a margin against zero (Appendix~\ref{app:corrections}).
The benchmark readings rest on agreement between two readers without an independent audit, and audits flag a few contrasts on the controlled tasks as sensitive to the reader (Appendix~\ref{app:readers}).
Finally, at $B=2$ the tests treat relabelled instances of three structural classes as separate problems, so their intervals do not generalise across classes (Appendix~\ref{app:stats}).

\section*{Acknowledgments}

The authors wish to acknowledge CSC -- IT Center for Science, Finland, for computational resources.

\bibliography{llm}

@article{geiping_scaling_2026,
	title = {Scaling up Test-Time Compute with Latent Reasoning: {A} Recurrent Depth Approach},
	volume = {38},
	shorttitle = {Scaling up {Test}-{Time} {Compute} with {Latent} {Reasoning}},
	url = {https://proceedings.neurips.cc/paper_files/paper/2025/hash/3b01972cf31e6fa0fe29e4b8b5c2a0a1-Abstract-Conference.html},
	doi = {10.52202/085713-1380},
	language = {en},
	urldate = {2026-10-12},
	journal = {Advances in Neural Information Processing Systems},
	author = {Geiping, Jonas and McLeish, Sean and Jain, Neel and Kirchenbauer, John and Singh, Siddharth and Bartoldson, Brian and Kailkhura, Bhavya and Bhatele, Abhinav and Goldstein, Tom},
	year = {2025},
	pages = {41340--41391},
}

@inproceedings{hao_training_2026,
	title = {Training Large Language Models to Reason in a Continuous Latent Space},
	author = {Hao, Shibo and Sukhbaatar, Sainbayar and Su, DiJia and Li, Xian and Hu, Zhiting and Weston, Jason and Tian, Yuandong},
	booktitle = {Second Conference on Language Modeling},
	year = {2025},
	url = {https://proceedings.colm.cc/paper_files/paper/2025/hash/ed032b08a8822c3635cdcd961012ce60-Abstract-Conference.html},
}

@misc{yang_qwen3_2025,
	title = {Qwen3 Technical Report},
	url = {http://arxiv.org/abs/2505.09388},
	doi = {10.48550/arXiv.2505.09388},
	urldate = {2026-10-12},
	publisher = {arXiv},
	author = {Yang, An and Li, Anfeng and Yang, Baosong and Zhang, Beichen and Hui, Binyuan and Zheng, Bo and Yu, Bowen and Gao, Chang and Huang, Chengen and Lv, Chenxu and Zheng, Chujie and Liu, Dayiheng and Zhou, Fan and Huang, Fei and Hu, Feng and Ge, Hao and Wei, Haoran and Lin, Huan and Tang, Jialong and Yang, Jian and Tu, Jianhong and Zhang, Jianwei and Yang, Jianxin and Yang, Jiaxi and Zhou, Jing and Zhou, Jingren and Lin, Junyang and Dang, Kai and Bao, Keqin and Yang, Kexin and Yu, Le and Deng, Lianghao and Li, Mei and Xue, Mingfeng and Li, Mingze and Zhang, Pei and Wang, Peng and Zhu, Qin and Men, Rui and Gao, Ruize and Liu, Shixuan and Luo, Shuang and Li, Tianhao and Tang, Tianyi and Yin, Wenbiao and Ren, Xingzhang and Wang, Xinyu and Zhang, Xinyu and Ren, Xuancheng and Fan, Yang and Su, Yang and Zhang, Yichang and Zhang, Yinger and Wan, Yu and Liu, Yuqiong and Wang, Zekun and Cui, Zeyu and Zhang, Zhenru and Zhou, Zhipeng and Qiu, Zihan},
	month = may,
	year = {2025},
	note = {{arXiv:2505.09388} [cs.CL]},
}

@article{jiang_artificial_2026,
	title = {Artificial Hivemind: The Open-Ended Homogeneity of Language Models (and Beyond)},
	volume = {38},
	shorttitle = {Artificial {Hivemind}},
	url = {https://proceedings.neurips.cc/paper_files/paper/2025/hash/754d5a526a5ee5a47220664a0eb92751-Abstract-Datasets_and_Benchmarks_Track.html},
	doi = {10.52202/085713-2732},
	language = {en},
	urldate = {2026-10-12},
	journal = {Advances in Neural Information Processing Systems},
	author = {Jiang, Liwei and Chai, Yuanjun and Li, Margaret and Liu, Mickel and Fok, Raymond and Dziri, Nouha and Tsvetkov, Yulia and Sap, Maarten and Albalak, Alon and Choi, Yejin},
	note = {NeurIPS 2025, Datasets and Benchmarks Track},
	year = {2025},
}

@misc{schaeffer_how_2026,
	title = {How Strong Is the Evidence for the Artificial Hivemind? {R}eevaluating Evidence for the Open-Ended Homogeneity of Language Models},
	shorttitle = {How {Strong} {Is} the {Evidence} for the {Artificial} {Hivemind}?},
	url = {https://arxiv.org/abs/2609.33936},
	doi = {10.48550/arXiv.2609.33936},
	language = {en},
	urldate = {2026-10-12},
	journal = {arXiv.org},
	author = {Schaeffer, Rylan and Miranda, Brando and Kazdan, Joshua and Chudnovsky, Jessica and Koyejo, Sanmi},
	month = sep,
	year = {2026},
	note = {{arXiv:2609.33936} [cs.LG]},
}

@inproceedings{wilie_belief_2024,
	address = {Miami, Florida, USA},
	title = {Belief Revision: The Adaptability of Large Language Models Reasoning},
	shorttitle = {Belief {Revision}},
	url = {https://aclanthology.org/2024.emnlp-main.586/},
	doi = {10.18653/v1/2024.emnlp-main.586},
	urldate = {2026-10-12},
	booktitle = {Proceedings of the 2024 {Conference} on {Empirical} {Methods} in {Natural} {Language} {Processing}},
	publisher = {Association for Computational Linguistics},
	author = {Wilie, Bryan and Cahyawijaya, Samuel and Ishii, Etsuko and He, Junxian and Fung, Pascale},
	editor = {Al-Onaizan, Yaser and Bansal, Mohit and Chen, Yun-Nung},
	month = nov,
	year = {2024},
	pages = {10480--10496},
}

@inproceedings{laban_llms_2025,
	title = {{LLMs} Get Lost In Multi-Turn Conversation},
	author = {Laban, Philippe and Hayashi, Hiroaki and Zhou, Yingbo and Neville, Jennifer},
	booktitle = {The Fourteenth International Conference on Learning Representations},
	year = {2026},
	url = {https://openreview.net/forum?id=VKGTGGcwl6},
}

@misc{mehta_when_2026,
	title = {When Agents Commit Too Soon: Diagnosing Premature Commitment in {LLM} Agents},
	shorttitle = {When {Agents} {Commit} {Too} {Soon}},
	url = {http://arxiv.org/abs/2606.22936},
	doi = {10.48550/arXiv.2606.22936},
	urldate = {2026-10-12},
	publisher = {arXiv},
	author = {Mehta, Aman},
	month = jun,
	year = {2026},
	note = {{arXiv:2606.22936} [cs.AI]},
}

@inproceedings{kirk_understanding_2024,
	title = {Understanding the Effects of {RLHF} on {LLM} Generalisation and Diversity},
	author = {Kirk, Robert and Mediratta, Ishita and Nalmpantis, Christoforos and Luketina, Jelena and Hambro, Eric and Grefenstette, Edward and Raileanu, Roberta},
	booktitle = {International Conference on Learning Representations},
	year = {2024},
	url = {https://proceedings.iclr.cc/paper_files/paper/2024/hash/5a68d05006d5b05dd9463dd9c0219db0-Abstract-Conference.html},
}

@inproceedings{holtzman_curious_2020,
	title = {The Curious Case of Neural Text Degeneration},
	author = {Holtzman, Ari and Buys, Jan and Du, Li and Forbes, Maxwell and Choi, Yejin},
	booktitle = {International Conference on Learning Representations},
	year = {2020},
	url = {https://iclr.cc/virtual_2020/poster_rygGQyrFvH.html},
}

@inproceedings{wang_self-consistency_2023,
	title = {Self-Consistency Improves Chain of Thought Reasoning in Language Models},
	author = {Wang, Xuezhi and Wei, Jason and Schuurmans, Dale and Le, Quoc and Chi, Ed H. and Narang, Sharan and Chowdhery, Aakanksha and Zhou, Denny},
	booktitle = {International Conference on Learning Representations},
	year = {2023},
	url = {https://openreview.net/forum?id=1PL1NIMMrw},
}

@article{yao_tree_2023,
	title = {Tree of Thoughts: Deliberate Problem Solving with Large Language Models},
	volume = {36},
	shorttitle = {Tree of {Thoughts}},
	url = {https://proceedings.neurips.cc/paper_files/paper/2023/hash/271db9922b8d1f4dd7aaef84ed5ac703-Abstract-Conference.html},
	doi = {10.52202/075280-0517},
	language = {en},
	urldate = {2026-10-12},
	journal = {Advances in Neural Information Processing Systems},
	author = {Yao, Shunyu and Yu, Dian and Zhao, Jeffrey and Shafran, Izhak and Griffiths, Tom and Cao, Yuan and Narasimhan, Karthik},
	month = dec,
	year = {2023},
	pages = {11809--11822},
}

@article{liu_lost_2024,
	address = {Cambridge, MA},
	title = {Lost in the Middle: How Language Models Use Long Contexts},
	volume = {12},
	shorttitle = {Lost in the {Middle}},
	url = {https://aclanthology.org/2024.tacl-1.9/},
	doi = {10.1162/tacl_a_00638},
	urldate = {2026-10-12},
	journal = {Transactions of the Association for Computational Linguistics},
	publisher = {MIT Press},
	author = {Liu, Nelson F. and Lin, Kevin and Hewitt, John and Paranjape, Ashwin and Bevilacqua, Michele and Petroni, Fabio and Liang, Percy},
	year = {2024},
	pages = {157--173},
}

@inproceedings{lin_zebralogic_2025,
	title = {{ZebraLogic}: On the Scaling Limits of {LLMs} for Logical Reasoning},
	author = {Lin, Bill Yuchen and Le Bras, Ronan and Richardson, Kyle and Sabharwal, Ashish and Poovendran, Radha and Clark, Peter and Choi, Yejin},
	booktitle = {Proceedings of the 42nd International Conference on Machine Learning},
	series = {Proceedings of Machine Learning Research},
	volume = {267},
	pages = {37889--37905},
	publisher = {PMLR},
	year = {2025},
	url = {https://proceedings.mlr.press/v267/lin25i.html},
}

@article{haralick_increasing_1980,
	title = {Increasing Tree Search Efficiency for Constraint Satisfaction Problems},
	author = {Haralick, Robert M. and Elliott, Gordon L.},
	journal = {Artificial Intelligence},
	volume = {14},
	number = {3},
	pages = {263--313},
	year = {1980},
}

@book{russell_artificial_2020,
	title = {Artificial Intelligence: A Modern Approach},
	author = {Russell, Stuart and Norvig, Peter},
	edition = {4},
	publisher = {Pearson},
	year = {2020},
}

@inproceedings{chen_premise_2024,
	title = {Premise Order Matters in Reasoning with Large Language Models},
	author = {Chen, Xinyun and Chi, Ryan Andrew and Wang, Xuezhi and Zhou, Denny},
	booktitle = {Proceedings of the 41st International Conference on Machine Learning},
	series = {Proceedings of Machine Learning Research},
	volume = {235},
	pages = {6596--6620},
	publisher = {PMLR},
	year = {2024},
	url = {https://proceedings.mlr.press/v235/chen24i.html},
}

@inproceedings{lu_fantastically_2022,
	title = {Fantastically Ordered Prompts and Where to Find Them: Overcoming Few-Shot Prompt Order Sensitivity},
	author = {Lu, Yao and Bartolo, Max and Moore, Alastair and Riedel, Sebastian and Stenetorp, Pontus},
	booktitle = {Proceedings of the 60th Annual Meeting of the Association for Computational Linguistics (Volume 1: Long Papers)},
	pages = {8086--8098},
	address = {Dublin, Ireland},
	publisher = {Association for Computational Linguistics},
	month = may,
	year = {2022},
	url = {https://aclanthology.org/2022.acl-long.556/},
	doi = {10.18653/v1/2022.acl-long.556},
}

@inproceedings{zheng_large_2024,
	title = {Large Language Models Are Not Robust Multiple Choice Selectors},
	author = {Zheng, Chujie and Zhou, Hao and Meng, Fandong and Zhou, Jie and Huang, Minlie},
	booktitle = {The Twelfth International Conference on Learning Representations},
	year = {2024},
	url = {https://proceedings.iclr.cc/paper_files/paper/2024/hash/54dd9e0cff6d9214e20d97eb2a3bae49-Abstract-Conference.html},
}

@inproceedings{pezeshkpour_large_2024,
	title = {Large Language Models Sensitivity to The Order of Options in Multiple-Choice Questions},
	author = {Pezeshkpour, Pouya and Hruschka, Estevam},
	booktitle = {Findings of the Association for Computational Linguistics: NAACL 2024},
	pages = {2006--2017},
	address = {Mexico City, Mexico},
	publisher = {Association for Computational Linguistics},
	month = jun,
	year = {2024},
	url = {https://aclanthology.org/2024.findings-naacl.130/},
	doi = {10.18653/v1/2024.findings-naacl.130},
}

@inproceedings{gandhi_stream_2024,
	title = {Stream of Search ({SoS}): Learning to Search in Language},
	author = {Gandhi, Kanishk and Lee, Denise H. and Grand, Gabriel and Liu, Muxin and Cheng, Winson and Sharma, Archit and Goodman, Noah D.},
	booktitle = {First Conference on Language Modeling},
	year = {2024},
	url = {https://openreview.net/forum?id=2cop2jmQVL},
}

@inproceedings{lehnert_beyond_2024,
	title = {Beyond {A*}: Better Planning with Transformers via Search Dynamics Bootstrapping},
	author = {Lehnert, Lucas and Sukhbaatar, Sainbayar and Su, DiJia and Zheng, Qinqing and McVay, Paul and Rabbat, Michael and Tian, Yuandong},
	booktitle = {First Conference on Language Modeling},
	year = {2024},
	url = {https://proceedings.colm.cc/paper_files/paper/2024/hash/c201b96c3e20d2e5d9f773b42ea2a30d-Abstract-Conference.html},
}

@inproceedings{turpin_language_2023,
	title = {Language Models Don't Always Say What They Think: Unfaithful Explanations in Chain-of-Thought Prompting},
	author = {Turpin, Miles and Michael, Julian and Perez, Ethan and Bowman, Samuel R.},
	booktitle = {Advances in Neural Information Processing Systems},
	volume = {36},
	year = {2023},
	url = {https://proceedings.neurips.cc/paper_files/paper/2023/hash/ed3fea9033a80fea1376299fa7863f4a-Abstract.html},
	doi = {10.52202/075280-3275},
}

@inproceedings{padmakumar_does_2024,
	title = {Does Writing with Language Models Reduce Content Diversity?},
	author = {Padmakumar, Vishakh and He, He},
	booktitle = {The Twelfth International Conference on Learning Representations},
	year = {2024},
	url = {https://proceedings.iclr.cc/paper_files/paper/2024/hash/02dec8877fb7c6aa9a79f81661baca7c-Abstract-Conference.html},
}

@inproceedings{hopkins_can_2023,
	title = {Can {LLMs} Generate Random Numbers? {E}valuating {LLM} Sampling in Controlled Domains},
	author = {Renda, Alex and Hopkins, Aspen and Carbin, Michael},
	booktitle = {Sampling and Optimization in Discrete Space Workshop at ICML 2023},
	year = {2023},
	url = {https://people.csail.mit.edu/renda/llm-sampling-paper},
}

@inproceedings{zhang_verbalized_2026,
	title = {Verbalized Sampling: How to Mitigate Mode Collapse and Unlock {LLM} Diversity},
	author = {Zhang, Jiayi and Yu, Simon and Chong, Derek and Sicilia, Anthony and Tomz, Michael and Manning, Christopher D. and Shi, Weiyan},
	booktitle = {Proceedings of the 43rd International Conference on Machine Learning},
	series = {Proceedings of Machine Learning Research},
	volume = {306},
	pages = {160073--160152},
	publisher = {PMLR},
	year = {2026},
	url = {https://proceedings.mlr.press/v306/zhang26hy.html},
}

@inproceedings{west_base_2025,
	title = {Base Models Beat Aligned Models at Randomness and Creativity},
	author = {West, Peter and Potts, Christopher},
	booktitle = {Second Conference on Language Modeling},
	url = {https://proceedings.colm.cc/paper_files/paper/2025/hash/f77b8a0f769a62fa2d68ddbc884afabb-Abstract-Conference.html},
	year = {2025},
	note = {\href{https://arxiv.org/abs/2505.00047}{arXiv:2505.00047}},
}

@article{holm_simple_1979,
	title = {A Simple Sequentially Rejective Multiple Test Procedure},
	author = {Holm, Sture},
	journal = {Scandinavian Journal of Statistics},
	volume = {6},
	number = {2},
	pages = {65--70},
	year = {1979},
}

@article{schuirmann_comparison_1987,
	title = {A Comparison of the Two One-Sided Tests Procedure and the Power Approach for Assessing the Equivalence of Average Bioavailability},
	author = {Schuirmann, Donald J.},
	journal = {Journal of Pharmacokinetics and Biopharmaceutics},
	volume = {15},
	number = {6},
	pages = {657--680},
	year = {1987},
}

@book{efron_introduction_1993,
	title = {An Introduction to the Bootstrap},
	author = {Efron, Bradley and Tibshirani, Robert J.},
	publisher = {Chapman \& Hall},
	year = {1993},
}

@article{hill_diversity_1973,
	title = {Diversity and Evenness: A Unifying Notation and Its Consequences},
	author = {Hill, Mark O.},
	journal = {Ecology},
	volume = {54},
	number = {2},
	pages = {427--432},
	year = {1973},
}

@misc{openai_gpt-oss_2025,
	title = {{gpt-oss}-120b \& {gpt-oss}-20b Model Card},
	author = {{OpenAI}},
	url = {https://arxiv.org/abs/2508.10925},
	doi = {10.48550/arXiv.2508.10925},
	year = {2025},
	note = {{arXiv:2508.10925} [cs.CL]},
}

@misc{olmo_olmo3_2025,
	title = {Olmo 3},
	author = {{Team Olmo}},
	howpublished = {Technical report, Allen Institute for AI},
	url = {https://arxiv.org/abs/2512.13961},
	doi = {10.48550/arXiv.2512.13961},
	year = {2025},
	note = {{arXiv:2512.13961} [cs.CL]},
}

@misc{gemma_team_gemma4_2026,
	title = {Gemma 4 Technical Report},
	author = {{Gemma Team}},
	url = {https://arxiv.org/abs/2607.02770},
	doi = {10.48550/arXiv.2607.02770},
	year = {2026},
	note = {{arXiv:2607.02770} [cs.CL]},
}

@misc{zheng_natural_2024,
	title = {{NATURAL} {PLAN}: Benchmarking {LLMs} on Natural Language Planning},
	author = {Zheng, Huaixiu Steven and Mishra, Swaroop and Zhang, Hugh and Chen, Xinyun and Chen, Minmin and Nova, Azade and Hou, Le and Cheng, Heng-Tze and Le, Quoc V. and Chi, Ed H. and Zhou, Denny},
	url = {https://arxiv.org/abs/2406.04520},
	doi = {10.48550/arXiv.2406.04520},
	year = {2024},
	note = {{arXiv:2406.04520} [cs.CL]},
}

@inproceedings{fan_nphardeval_2024,
	title = {{NPHardEval}: Dynamic Benchmark on Reasoning Ability of Large Language Models via Complexity Classes},
	author = {Fan, Lizhou and Hua, Wenyue and Li, Lingyao and Ling, Haoyang and Zhang, Yongfeng},
	booktitle = {Proceedings of the 62nd Annual Meeting of the Association for Computational Linguistics (Volume 1: Long Papers)},
	pages = {4092--4114},
	address = {Bangkok, Thailand},
	publisher = {Association for Computational Linguistics},
	doi = {10.18653/v1/2024.acl-long.225},
	year = {2024},
}

@inproceedings{kwon_efficient_2023,
	title = {Efficient Memory Management for Large Language Model Serving with {PagedAttention}},
	author = {Kwon, Woosuk and Li, Zhuohan and Zhuang, Siyuan and Sheng, Ying and Zheng, Lianmin and Yu, Cody Hao and Gonzalez, Joseph E. and Zhang, Hao and Stoica, Ion},
	booktitle = {Proceedings of the 29th Symposium on Operating Systems Principles},
	year = {2023},
}

@inproceedings{gekhman_inside-out_2025,
	title = {Inside-Out: Hidden Factual Knowledge in {LLMs}},
	author = {Gekhman, Zorik and Ben-David, Eyal and Orgad, Hadas and Ofek, Eran and Belinkov, Yonatan and Szpektor, Idan and Herzig, Jonathan and Reichart, Roi},
	booktitle = {Second Conference on Language Modeling},
	year = {2025},
	url = {https://proceedings.colm.cc/paper_files/paper/2025/hash/90d757586e36d0f5f227660c5a6bc782-Abstract-Conference.html},
}

@inproceedings{yue_does_2025,
	title = {Does Reinforcement Learning Really Incentivize Reasoning Capacity in {LLMs} Beyond the Base Model?},
	author = {Yue, Yang and Chen, Zhiqi and Lu, Rui and Zhao, Andrew and Wang, Zhaokai and Yue, Yang and Song, Shiji and Huang, Gao},
	booktitle = {Advances in {Neural} {Information} {Processing} {Systems}},
	volume = {38},
	year = {2025},
	url = {https://proceedings.neurips.cc/paper_files/paper/2025/hash/537d5aa768c2d534016a4d06f87bc8fb-Abstract-Conference.html},
	doi = {10.52202/085713-1933},
}

@inproceedings{puri_reaching_2026,
	title = {Escaping the Mode: Multi-Answer Reinforcement Learning in {LMs}},
	author = {Puri, Isha and Damani, Mehul and Shenfeld, Idan and Ghassemi, Marzyeh and Andreas, Jacob and Kim, Yoon},
	booktitle = {Proceedings of the 43rd International Conference on Machine Learning},
	series = {Proceedings of Machine Learning Research},
	volume = {306},
	pages = {100127--100146},
	publisher = {PMLR},
	url = {https://proceedings.mlr.press/v306/puri26a.html},
	year = {2026},
	note = {\href{https://arxiv.org/abs/2603.24844}{arXiv:2603.24844}},
}

@inproceedings{zhang_noveltybench_2025,
	title = {{NoveltyBench}: Evaluating Language Models for Humanlike Diversity},
	author = {Zhang, Yiming and Diddee, Harshita and Holm, Susan and Liu, Hanchen and Liu, Xinyue and Samuel, Vinay and Wang, Barry and Ippolito, Daphne},
	booktitle = {Second Conference on Language Modeling},
	year = {2025},
	url = {https://proceedings.colm.cc/paper_files/paper/2025/hash/a6a066fb44f2fe0d36cf740c873b8890-Abstract-Conference.html},
}

@inproceedings{zhang_forcing_2024,
	title = {Forcing Diffuse Distributions out of Language Models},
	author = {Zhang, Yiming and Schwarzschild, Avi and Carlini, Nicholas and Kolter, Zico and Ippolito, Daphne},
	booktitle = {First Conference on Language Modeling},
	year = {2024},
	url = {https://openreview.net/forum?id=9JY1QLVFPZ},
}

@misc{wang_investigating_2026,
	title = {Investigating Human--{AI} Discrepancies via Multiple-Solution Problems},
	author = {Wang, Zihao and Insulla, Francesco and Montanari, Andrea},
	url = {https://arxiv.org/abs/2609.34258},
	doi = {10.48550/arXiv.2609.34258},
	year = {2026},
	note = {{arXiv:2609.34258} [cs.CY]},
}

@inproceedings{daliberti_measuring_2026,
	title = {Measuring and Mitigating Mode Collapse in {RLVR}},
	author = {d'Aliberti, Liv G. and Abdulhai, Marwa and Druchyna, Sofiia and Henderson, Peter and Horta Ribeiro, Manoel},
	booktitle = {6th Workshop on Mathematical Reasoning and {AI} at {NeurIPS} 2026},
	url = {https://openreview.net/forum?id=YoYMNRri3W},
	year = {2026},
	note = {\href{https://arxiv.org/abs/2610.11064}{arXiv:2610.11064}},
}

@inproceedings{yun_price_2025,
	title = {The Price of Format: Diversity Collapse in {LLM}s},
	author = {Yun, Longfei and An, Chenyang and Wang, Zilong and Peng, Letian and Shang, Jingbo},
	editor = {Christodoulopoulos, Christos and Chakraborty, Tanmoy and Rose, Carolyn and Peng, Violet},
	booktitle = {Findings of the {A}ssociation for {C}omputational {L}inguistics: {EMNLP} 2025},
	month = nov,
	year = {2025},
	address = {Suzhou, China},
	publisher = {Association for Computational Linguistics},
	url = {https://aclanthology.org/2025.findings-emnlp.836/},
	doi = {10.18653/v1/2025.findings-emnlp.836},
	pages = {15454--15468},
}

@misc{karouzos_where_2026,
	title = {Where Does Output Diversity Collapse in Post-Training?},
	author = {Karouzos, Constantinos and Tan, Xingwei and Aletras, Nikolaos},
	url = {https://arxiv.org/abs/2604.16027},
	year = {2026},
	note = {{arXiv:2604.16027} [cs.CL]},
}

@misc{banayeeanzade_sampling_2026,
	title = {Sampling More, Getting Less: Calibration is the Diversity Bottleneck in {LLMs}},
	author = {Banayeeanzade, Amin and Yang, Qingchuan and Tarsadiya, Dhruv and Bahrani, Fatemeh and Blas, Leonardo and Samuel, Alfy and Jia, Robin and Razaviyayn, Meisam and Karimireddy, Sai Praneeth},
	url = {https://arxiv.org/abs/2605.11128},
	year = {2026},
	note = {{arXiv:2605.11128} [cs.CL]},
}

@misc{guan_order_2025,
	title = {The Order Effect: Investigating Prompt Sensitivity to Input Order in {LLMs}},
	author = {Guan, Bryan and Roosta, Tanya and Passban, Peyman and Rezagholizadeh, Mehdi},
	url = {https://arxiv.org/abs/2502.04134},
	year = {2025},
	note = {{arXiv:2502.04134} [cs.CL]},
}

@misc{parikh_one-word_2026,
	title = {The One-Word Census: Answer-Choice Conformity Across 105 Language Models},
	author = {Parikh, Tapan},
	url = {https://arxiv.org/abs/2607.12796},
	year = {2026},
	note = {{arXiv:2607.12796} [cs.CL]},
}

@inproceedings{gomes_boosting_1998,
	title = {Boosting Combinatorial Search Through Randomization},
	author = {Gomes, Carla P. and Selman, Bart and Kautz, Henry},
	booktitle = {Proceedings of the Fifteenth National Conference on Artificial Intelligence},
	pages = {431--437},
	year = {1998},
	publisher = {AAAI Press},
	address = {Madison, Wisconsin},
}

@misc{sharma_schedbench_2026,
	title = {{SCHEDBench}: A Benchmark for Evaluating {LLM} Constraint Faithfulness in Natural-Language Combinatorial Scheduling},
	author = {Sharma, Shrenil Shaun and Sharma, Avi},
	url = {https://arxiv.org/abs/2608.00991},
	year = {2026},
	note = {{arXiv:2608.00991} [cs.AI]},
}

@inproceedings{jinnai_generating_2024,
	title = {Generating Diverse and High-Quality Texts by Minimum {B}ayes Risk Decoding},
	author = {Jinnai, Yuu and Honda, Ukyo and Morimura, Tetsuro and Zhang, Peinan},
	booktitle = {Findings of the Association for Computational Linguistics: ACL 2024},
	pages = {8494--8525},
	publisher = {Association for Computational Linguistics},
	address = {Bangkok, Thailand},
	year = {2024},
	url = {https://aclanthology.org/2024.findings-acl.503},
}

\clearpage
\appendix
\setcounter{figure}{0}
\setcounter{table}{0}
\renewcommand{\thefigure}{A\arabic{figure}}
\renewcommand{\thetable}{A\arabic{table}}
\renewcommand{\topfraction}{0.9}
\renewcommand{\dbltopfraction}{0.9}
\renewcommand{\textfraction}{0.08}
\renewcommand{\floatpagefraction}{0.7}
\renewcommand{\dblfloatpagefraction}{0.7}
\setcounter{topnumber}{3}
\setcounter{dbltopnumber}{3}
\setcounter{totalnumber}{5}
\raggedbottom

\section{Tasks and conditions}
\label{app:tasks}

\paragraph{Symbolic problems.}
We enumerated sets of $B\in\{2,\dots,6\}$ assignments of five entities to five values, retained those whose allowed entity--value pairs yield exactly those assignments, and grouped the resulting problems into structural classes that are equivalent under renaming entities and values.
For $B=2$ the main set uses eight differently labelled instances of each of three classes in which the two solutions differ in at least three entities.
For $B=4$ and $B=6$ it uses one instance from each of 24 classes, out of the 27 and 66 classes in the census.
The exclusion facts of a problem exclude exactly the entity--value pairs that occur in none of its solutions, so every pair still allowed occurs in at least one solution.
A problem was shown in its canonical form, with relabelled values, with its facts listed in reverse entity order, or with both changes.
Every answer is mapped back to the canonical solution set before scoring.
The free-generation prompt for the first $B=2$ problem reads in full:

\begin{quote}\small\ttfamily
Each of E1, E2, E3, E4, E5 occupies a different value from V1, V2, V3, V4, V5.\\
E1 is not in V2.\ E1 is not in V4.\ E1 is not in V5.\
E2 is not in V2.\ E2 is not in V3.\ E2 is not in V4.\
E3 is not in V1.\ E3 is not in V3.\ E3 is not in V4.\
E4 is not in V1.\ E4 is not in V2.\ E4 is not in V3.\ E4 is not in V5.\
E5 is not in V1.\ E5 is not in V4.\ E5 is not in V5.\\[2pt]
No additional fact is supplied.\\
Any complete world satisfying the original facts is acceptable.\\
Question: What value does E2 have?\\
Reconstruct the complete world.\\[2pt]
You may explain your reasoning before answering. Finish with one clearly marked final answer in this form:\\
FINAL: E1=<value>, E2=<value>, E3=<value>, E4=<value>, E5=<value>; consequence=<value>\\
Give the complete assignment and the answer to the question. Do not give multiple alternative final answers.
\end{quote}

\noindent (Each fact is on its own line in the prompt.)
The two solutions are $(E_1,\dots,E_5)=(V_1,V_5,V_2,V_4,V_3)$ and $(V_3,V_1,V_5,V_4,V_2)$.
A cued prompt replaces the two lines ``No additional fact is supplied.'' and ``Any complete world \dots'' with one line, ``Real additional fact: E1 is in V1.'' for the first solution or ``Real additional fact: E1 is in V3.'' for the second, and is otherwise identical.
Here the fact fixes one entity, and the model must still derive the other four.
In general a cue is the smallest set of assignments, none of them for the queried entity, that leaves only the target, and several assignments share the line (``Real additional fact: E1 is in V3. E4 is in V1.'').
Of the 288 Symbolic targets, 197 are cued by one assignment, 84 by two and 7 by three; at $B=6$, half of the 144 targets need more than one.
Each ZebraLogic target is cued by one added clue.

The \texttt{consequence} field asks for the answer to the question, the value of the queried entity (here $E_2$).
In free generation, validity depends only on the complete assignment: a stated consequence that disagrees with the assignment does not make the output invalid.
Cued success additionally requires that the consequence, if stated, equals the target's value for the queried entity.

\paragraph{ZebraLogic variants.}
From 48 source puzzles of ZebraLogic~\citep{lin_zebralogic_2025} (the \texttt{grid\_mode} test split of \texttt{allenai/ZebraLogicBench} at revision \texttt{2f94a44}), we removed clues until $B=2$, 3 or 4 solutions remained (24, 8 and 16 variants; 12 for each of the sizes $3\times3$, $3\times4$, $4\times3$ and $4\times4$).
Pilot puzzles were disjoint from these.
Each of the 136 solutions has a cue: one additional clue that leaves only that solution.
Every displayed prompt was re-parsed and solved, and its solution set was required to match the intended one.

The free-generation prompt for the $3\times3$, $B=2$ variant of source puzzle \texttt{lgp-test-3x3-5}, from which one clue was removed, reads:

\begin{quote}\small\ttfamily
There are 3 houses, numbered 1 to 3 from left to right, as seen from across the street. Each house is occupied by a different person. Each house has a unique attribute for each of the following characteristics:\\
\hspace*{1em}- Each person has a unique name: `Eric`, `Peter`, `Arnold`\\
\hspace*{1em}- The people are of nationalities: `dane`, `brit`, `swede`\\
\hspace*{1em}- People have unique favorite sports: `basketball`, `tennis`, `soccer`\\[2pt]
\#\# Clues:\\
1. Eric is in the second house.\\
2. The Dane is somewhere to the right of Arnold.\\
3. The British person is somewhere to the right of the person who loves tennis.\\
4. The British person is not in the third house.\\[2pt]
Some clues have been removed, so more than one arrangement may satisfy every clue. Any arrangement that satisfies every clue is acceptable.\\[2pt]
Finish with one clearly marked final grid line in this form:\\
FINAL: House 1: <Attribute>=<value>, \dots; House 2: \dots
\end{quote}

\noindent The two solutions agree on Arnold (Swede, tennis) in house 1, Eric (Brit) in house 2 and Peter (Dane) in house 3, and differ only in whether Eric or Peter loves soccer.
Each cued prompt adds one line before the answer instruction, for example ``Real additional clue: The person who loves soccer is in the third house.'' (or ``\dots{} basketball \dots'' for the other solution).

\paragraph{Intervention prompts.}
The examples quote the prompts shown to the models.
\emph{Randomised presentation} applies a random value relabelling and a random order in which the entities' facts are listed, drawn independently for each request; on ZebraLogic it independently permutes each attribute's declared value list, the order of the attribute lines and the order of the clues (renumbered), and leaves house numbering and clue wording unchanged.
Answers are mapped back before scoring.
The Symbolic \emph{random-choice instruction} adds, after ``Any complete world satisfying the original facts is acceptable.'', the line ``If several complete worlds satisfy the facts, choose one of them uniformly at random.''; on ZebraLogic it is ``If several arrangements satisfy every clue, choose one of them uniformly at random.''
\emph{Enumerate-then-choose} replaces the question with ``Enumerate every complete world satisfying the original facts, one complete world per line. Then choose one admissible world uniformly at random and give it as the single FINAL answer.''
The Symbolic \emph{specified-order} instruction is inserted at the same place, for example:

\begin{quote}\small\ttfamily
Solve the problem by considering the entities in this order: E3, E1, E4, E2, E5.\\
For each entity, try its possible values in this order: V2, V1, V3, V5, V4.\\
Keep the first choice that leads to a complete world, and give that world.
\end{quote}

\noindent The entity-only and value-only variants keep the first or second sentence and the last one.
For the fixed orders, the entity and value orders are the first permutations in a fixed hash order whose first solution differs from both the MRV-like and the name-order prediction.
For the \emph{random search order}, we draw the entity order and the value order independently and uniformly from all $120\times120$ pairs for each request and freeze them before generation.
On ZebraLogic the order instruction follows the sentence explaining that several arrangements are acceptable; one prompt reads as follows.

\begin{quote}\small\ttfamily
Fill the grid house by house, from House 1 to House 3. Within each house, decide the attributes in this order: FavoriteSport, Nationality, Name. For each attribute, try its values in this order: FavoriteSport: soccer, basketball, tennis; Nationality: dane, swede, brit; Name: Arnold, Peter, Eric. Keep the first choice that leads to an arrangement satisfying every clue.
\end{quote}

\noindent For the fixed order, the attribute order and each attribute's value order are chosen by a fixed hash rule so that their first solution differs from the plain house-by-house prediction; for the random order they are drawn for each request.

\section{Models and decoding}
\label{app:models}

Models were served with vLLM 0.19.1~\citep{kwon_efficient_2023} from pinned Hugging Face revisions, using each model's native chat template, with the settings in Table~\ref{tab:decoding}.
We added no system message; templates that supply a default one keep it, so templates and default system messages differ between models and are the same in every condition of a model.
The gpt-oss template writes the current date into its system message; the date is the same for every request of an experiment.
The gpt-oss models use their native reasoning format at medium reasoning effort, and both Qwen3 models and Gemma run with thinking enabled; Gemma's template adds a system turn that carries its thinking control token.
The Olmo~3 7B SFT template supplies a default assistant system message, whereas the DPO and RLVR templates share a different one that describes function calling with an empty function list; the user text, sampling settings and thinking prefix are identical.

\begin{table*}[t]
  \centering\footnotesize
  \setlength{\tabcolsep}{5pt}
  \begin{tabular}{@{}lllrr@{}}
    \toprule
    Model & Parameters & Sampling & First allowance & Retry allowance \\
    \midrule
    Qwen3-1.7B (thinking) & 1.7B & $T{=}0.6$, top-$p$ 0.95, top-$k$ 20 & 16,384 & 32,768 \\
    Olmo-3.1-32B-Instruct & 32.2B$^{a}$ & $T{=}1.0$, top-$p$ 1.0 & 8,192 & 16,384 \\
    gpt-oss-20b & 21B (3.6B) & $T{=}1.0$, top-$p$ 1.0, effort medium & 16,384 & 32,768 \\
    gpt-oss-120b & 117B (5.1B) & $T{=}1.0$, top-$p$ 1.0, effort medium & 16,384 & 32,768 \\
    Qwen3-30B-A3B (thinking) & 30.5B (3.3B) & $T{=}0.6$, top-$p$ 0.95, top-$k$ 20 & 16,384 & 32,768 \\
    Gemma-4-26B-A4B-it & 25.2B (3.8B) & $T{=}1.0$, top-$p$ 0.95, top-$k$ 64 & 16,384 & 32,768 \\
    Olmo 3 7B Think, three stages & 7.3B$^{a}$ & $T{=}0.6$, top-$p$ 0.95 & 32,768 & $65{,}536-\text{prompt}$ \\
    \bottomrule
  \end{tabular}
  \caption{\textbf{Models, sampling settings and new-token allowances} on the Symbolic and ZebraLogic tasks.
  Parameter counts are those stated on the model cards; for the four mixture-of-experts models the parameters active per token follow in parentheses.
  $^{a}$The model card states no total; this is the total of the released safetensors weights.
  All models except Olmo-3.1-32B-Instruct use the vendor-recommended sampling settings; Olmo-3.1-32B-Instruct was sampled at $T{=}1.0$ and top-$p$ 1.0 instead of its recommended $T{=}0.6$ and top-$p$ 0.95.
  The answer-only condition uses 48 new tokens without retry for every model; the allowances on the public benchmarks were set by their pilots (Appendix~\ref{app:benchmarks}).}
  \label{tab:decoding}
\end{table*}

\paragraph{OLMo at its recommended decoding.}
Because OLMo was sampled at $T{=}1.0$ and top-$p$ 1.0, we repeated four of its pre-specified tests at its recommended $T{=}0.6$ and top-$p$ 0.95.
All four pass: the default's excess over the uniform expectation (42.4 points), the MRV-like minus name-order share on the held-out presentations (45.4 points), and the breadth gains under randomised presentation on the Symbolic task (45.3 points) and on ZebraLogic (21.1 points), each with validity within its margin (Table~\ref{tab:secondary} gives the intervals).

\paragraph{Allowances and retries.}
If a response stops at its allowance without an end-of-text token, the same request is retried once with the same random seed at the retry allowance; we keep the retry's response, never the better-scoring of the two.
Responses still truncated count as invalid.
The Olmo~3 7B retry allowance is computed per request as the 65,536-token context minus the prompt length.

\paragraph{Sample counts.}
\emph{Free generation and every comparison condition:} the first 32 executed requests per problem and condition (Symbolic: 72 problems, 2,304 requests per condition; ZebraLogic: 48 variants, 1,536).
\emph{Cued generation:} 8 requests per target solution for the main models and gpt-oss-120b (Symbolic: 288 targets; ZebraLogic: 136 targets) and 4 per target for the Olmo~3 7B training stages.
\emph{Prospective test:} 48 problems $\times$ 4 presentations $\times$ 32 requests, 6,144 per model.
\emph{Order manipulations:} 32 requests per problem and condition on the 48 problems with $B=4$ or 6, and on all 72 problems for the random search order; the instruction that fixes both orders uses 47 problems, because for one problem no value order met the selection rule of Appendix~\ref{app:tasks} and it was excluded before generation.
\emph{New models and benchmarks:} the two new models used the same counts on the Symbolic and ZebraLogic tasks; the public benchmarks use 32 requests per instance and condition and 8 per cue target (Appendix~\ref{app:benchmarks}).
\emph{Exceptions:} Qwen3-1.7B's Symbolic rank profiles (Figures~\ref{fig:concentration}C and~\ref{fig:appendix-rank}) use 48 free requests per problem and 16 cued requests per target, and OLMo's history experiment uses 32 free and 16 cued follow-up requests per problem and target (Appendix~\ref{app:history}).

\paragraph{Compute and licences.}
All generation ran on single NVIDIA GH200 GPUs, at most 16 at a time, and used about 860 GPU-hours in total for the experiments reported here, including pilots, failed starts and retries.
The weights of all models are released under the Apache-2.0 licence.
NATURAL PLAN releases its software under the Apache-2.0 licence and all other materials under CC~BY~4.0; NPHardEval releases its source code under the Apache-2.0 licence, states that it is intended solely for research use and gives no separate licence for its data; the ZebraLogicBench dataset card declares no licence.
We release our derived instances under the terms of their sources, with attribution.

\section{Statistical analysis}
\label{app:stats}

\newcommand{\rh}[1]{\rlap{\rotatebox{55}{#1}}}
\newcommand{\testgroup}[1]{\multicolumn{9}{@{}l}{\textit{#1}} \\}
\begin{table*}[t]
  \centering\footnotesize
  \setlength{\tabcolsep}{3.5pt}
  \begin{tabular}{@{}>{\raggedright\arraybackslash}p{4.15cm}>{\raggedright\arraybackslash}p{6.3cm}cccccc@{\hspace{9pt}}l@{}}
    \toprule
    Test & Passes when & \rh{Qwen3-1.7B} & \rh{OLMo} & \rh{gpt-oss-20b} & \rh{gpt-oss-120b} & \rh{Qwen3-30B-A3B} & \rh{Gemma} & Sec. \\
    \midrule
    \testgroup{Concentration and shared defaults}
    Concentration, Symbolic & the default's share of valid outputs exceeds uniform sampling's, over 72 problems & & & & & \yes & \yes & \ref{sec:concentration-results} \\
    Concentration, ZebraLogic & the same, over 48 puzzles & \yes & \yes & \yes & \yes & \yes & \yes & \ref{sec:concentration-results} \\
    Shared defaults, pairs with a new model & two models' defaults coincide more often than under uniform relabelling, on both testbeds & \multicolumn{6}{c}{all nine pairs pass} & \ref{app:shared-defaults} \\
    Shared defaults at scale & gpt-oss-120b's defaults agree with gpt-oss-20b's more often than with its defaults on other problems & & & & \yes & & & \ref{app:figure-notes} \\
    \addlinespace[3pt]
    \testgroup{Recoverability}
    Cued success by rank, Symbolic & its slope over name-order rank lies within $\pm5$ points per rank & & \yes & \yes & & & & \ref{sec:recoverability} \\
    Cued success, Symbolic & non-default minus default cued success lies above $-10$ points & & & & & \yes & \yes & \ref{sec:recoverability} \\
    Cued success, ZebraLogic & the same & \yes & \yes & \yes & \yes & \yes & \yes & \ref{sec:recoverability} \\
    History & an earlier answer raises its own free share, and cued success for the other solutions stays within $\pm5$ points & & \yes & & & & & \ref{app:history} \\
    \addlinespace[3pt]
    \testgroup{Ordered search}
    Held-out prediction & more requests match the MRV-like than the name-order prediction & \yes & \yes & \yes & \yes$^{a}$ & \yes & \yes & \ref{sec:predict} \\
    Own procedure at scale & gpt-oss-120b matches its procedure more often than gpt-oss-20b matches its own & & & & \yes & & & \ref{app:stages} \\
    Candidates or mentions, restated facts & outputs follow candidate count, not mention count (two-sided) & \yes & \yes & \yes & \yes & & & \ref{sec:predict} \\
    Candidates or mentions, disjunctive fact & the same (two-sided) & \no & \no & \yes & \yes & & & \ref{sec:predict} \\
    Reversed orders, Symbolic & reversing the declared and exclusion-listing orders moves outputs to the new prediction & \yes & \yes & \yes & & & & \ref{sec:declaration} \\
    Order factorial, Symbolic & declaration order and, separately, exclusion-listing order move the predicted solution's share & \yes & \yes & \yes & & & & \ref{sec:declaration} \\
    Reversed value order, ZebraLogic & reversing each attribute's value order moves outputs to the new prediction & \yes & \yes & \yes & \yes & \yes & \yes & \ref{sec:declaration} \\
    Specified order, Symbolic & the instruction raises the share of the solution it specifies (both orders; main models also each order alone) & \yes & \yes & \yes & & \yes & \yes & \ref{sec:specify} \\
    Specified order, ZebraLogic & the same, for the house, attribute and value order & \yes & \yes & \yes & & \yes & \yes & \ref{sec:specify} \\
    \bottomrule
  \end{tabular}
  \caption{\textbf{Pre-specified tests on concentration, recoverability and ordered search}, with their outcomes after Holm correction within each model's family of tests in each study.
  \yes: passed; \no: not passed; blank: no such test for this model.
  Sec.: where the result is reported.
  $^{a}$Per presentation.}
  \label{tab:prespecified}
\end{table*}

\begin{table*}[t]
  \centering\footnotesize
  \setlength{\tabcolsep}{3.5pt}
  \begin{tabular}{@{}>{\raggedright\arraybackslash}p{4.15cm}>{\raggedright\arraybackslash}p{6.3cm}cccccc@{\hspace{9pt}}l@{}}
    \toprule
    Test & Passes when & \rh{Qwen3-1.7B} & \rh{OLMo} & \rh{gpt-oss-20b} & \rh{gpt-oss-120b} & \rh{Qwen3-30B-A3B} & \rh{Gemma} & Sec. \\
    \midrule
    \testgroup{Interventions on the Symbolic task}
    Randomised presentation & breadth rises and the validity change lies above $-3$ points & \yes & \yes & \yes & \yes & \yes & \yes & \ref{sec:symbolic-interventions} \\
    \quad with renamed entities & the same & \yes & \yes$^{\dagger}$ & \yes & & & & \ref{sec:symbolic-interventions} \\
    \quad renaming adds breadth & breadth is higher with renamed entities than without & \no & \yes$^{b}$ & \yes & & & & \ref{app:readers} \\
    \quad against random choice & it broadens more than the random-choice instruction & \yes & \yes & \yes & & \yes & \yes & \ref{sec:symbolic-interventions} \\
    \quad against Verbalized Sampling & it broadens more (two-sided) & \yes & \yes & \yes & -- & & & \ref{app:baselines} \\
    \quad against the validity-preserving temperature & it broadens more & \yes & \yes & \yes & -- & & & \ref{app:baselines} \\
    Random search order & breadth rises and the validity change lies above $-3$ points & \yes & \yes$^{\dagger}$ & \yes & & \yes & \yes & \ref{sec:symbolic-interventions} \\
    Value labels alone & the same & \yes & \yes & \yes & & & & \ref{app:pa} \\
    Entity names or fact order & the one the model's tie rule forecasts broadens more & \no & \yes & \yes & & & & \ref{app:pa} \\
    Temperature $+0.5$ & validity falls & \no & \yes & \yes & & & & \ref{app:temperature} \\
    \addlinespace[3pt]
    \testgroup{Interventions on ZebraLogic}
    Randomised presentation & breadth rises and the validity change lies above $-5$ points & \no & \yes & \yes & \yes & \yes & \yes & \ref{sec:zebra-interventions} \\
    \quad against random choice & it broadens more than the random-choice instruction & \yes & \yes & \yes & & \yes & \yes & \ref{sec:zebra-interventions} \\
    \quad against Verbalized Sampling & it broadens more (two-sided) & \yes & \yes & \no & \no & & & \ref{app:baselines} \\
    \quad against the validity-preserving temperature & it broadens more & \yes & \yes & -- & \yes & & & \ref{app:baselines} \\
    Random search order & breadth rises and the validity change lies above $-5$ points & \yes$^{\dagger}$ & \yes & \yes & & \yes & \yes & \ref{sec:zebra-interventions} \\
    \addlinespace[3pt]
    \testgroup{Other protocols}
    Answer-only: validity & validity falls without written reasoning & \yes & \yes & \yes & \yes & & & \ref{app:answer-only} \\
    Answer-only: held-out contrast & the MRV-like minus name-order contrast weakens & \no$^{c}$ & \yes & \yes & \yes & & & \ref{app:answer-only} \\
    Reduced reasoning: concentration & the default's share still exceeds uniform sampling's & \yes & & \yes & \yes & & & \ref{app:answer-only} \\
    Reduced reasoning: held-out contrast & more requests still match the MRV-like prediction & \yes & & \yes & \yes & & & \ref{app:answer-only} \\
    Recommended decoding (OLMo) & concentration, the held-out contrast and randomised presentation on both tasks pass again & \multicolumn{6}{c}{all four pass} & \ref{app:models} \\
    Another solution, Symbolic & each further answer is the next solution of the model's ordered search more often than a uniform choice & \no & \no & \yes & \yes & & & \ref{app:it} \\
    Another solution, ZebraLogic & the same & \yes & \no & \no & \no & & & \ref{app:it} \\
    \bottomrule
  \end{tabular}
  \caption{\textbf{Pre-specified tests on the interventions and on other protocols.}
  Symbols as in Table~\ref{tab:prespecified}; --: undefined.
  Validity criteria are one-sided: the lower limit of the 90\% interval of the change must lie above the stated bound.
  $^{\dagger}$The two-sided equivalence test originally specified did not pass, because validity rose beyond the upper margin.
  $^{b}$Reader-sensitive (Appendix~\ref{app:readers}). $^{c}$Six problems had enough valid answers.
  The ten tests on the Olmo~3 7B Think training stages are reported in Appendix~\ref{app:stages}, the public-benchmark tests in Table~\ref{tab:benchmark-estimates}, and the ZebraLogic tests of the original runs in Table~\ref{tab:readers}.}
  \label{tab:prespecified-b}
\end{table*}

Intervals are percentile intervals from 2,000 bootstrap resamples~\citep{efron_introduction_1993} of clusters of related problems.
On the Symbolic task, a cluster is a structural class (problems equivalent under renaming entities and values) at $B=4$ and $B=6$, where each class occurs once in a problem set.
At $B=2$, the 24 problems are eight relabelled instances of each of three classes.
The pre-specified tests of all six models, and the other analyses of the four earlier models, treat each of these problems as its own cluster.
Two sets of analyses resample the three classes instead, giving 51 clusters over the 72 problems and three at $B=2$: the shared-default comparisons of all 15 pairs (Appendix~\ref{app:shared-defaults}), and the descriptive estimates of the two new models, namely their $B=2$ points in Figure~\ref{fig:concentration}A and their random-choice comparison in Figure~\ref{fig:interventions}.
The cluster counts of the tests are therefore:
72 (24 classes at each of $B=4$ and 6, plus 24 problems at $B=2$) for analyses on all 72 problems, including concentration, gpt-oss-120b's agreement with gpt-oss-20b, cued success, the interventions and the training stages;
48 classes for analyses on the $B=4$ and 6 problems only, namely the declared-order and exclusion-listing-order effects and the specified orders (47 for both orders together);
48 classes for the prospective test, with the four presentations of a problem resampled together;
46 and 43 problems for the history comparison;
and 48 source puzzles, one variant each, on ZebraLogic.
Intervals at $B=4$ and 6 therefore generalise over structural classes like those sampled, whereas at $B=2$ the per-problem intervals describe relabelled instances of only three classes and do not generalise across classes, and intervals that resample the three classes rest on too few clusters to be reliable.
On the $B=4$ and 6 problems alone, every Symbolic estimate whose interval excludes zero on all 72 problems keeps its direction and an interval excluding zero, with two exceptions among the descriptive comparisons: OLMo's validity drops under the random-choice instruction and the higher temperature (1.7 and 2.1 points) become 0.8 and 1.3 points, with intervals including zero (Table~\ref{tab:secondary}; for these two drops the stored interval is a 90\% interval and the one on the subset a 95\% interval).
These two estimates cover 60 of the 72 problems; on all 72, as in Figure~\ref{fig:interventions}, both drops are 2.3 points.
Conversely, the changes in cued success between the Olmo~3 training stages, both below one point, have intervals excluding zero on these problems (Table~\ref{tab:secondary}).

Validity changes are tested for non-inferiority: before the multiplicity correction described below, a change passes when the lower limit of its two-sided 90\% interval lies above minus the pre-specified margin, 3 points for Symbolic validity and 5 points for ZebraLogic validity (Appendix~\ref{app:benchmarks} gives the benchmark margins).
Cued success of non-default versus default solutions has a 10-point margin, and the slope of cued success over rank $\pm5$ points per rank step.
This one-sided rule was pre-specified for sets 5 and 6 of Table~\ref{tab:studies}.
Sets 1 to 4 had specified two-sided equivalence tests~\citep{schuirmann_comparison_1987}, which also fail when validity rises beyond the upper margin; for them the one-sided rule is post hoc, applied within the original Holm families, and the two-sided decision is reported beside it.
Four decisions differ, all from not supported to supported: OLMo's Symbolic random search order and its randomised presentation with renamed entities, gpt-oss-20b's random search order in the original ZebraLogic runs, and Qwen3-1.7B's random search order in the ZebraLogic re-run.
In the three that Figures~\ref{fig:interventions} and~\ref{fig:appendix-interventions} draw, validity rose, and its interval extends beyond the upper margin (3 points on the Symbolic task, 5 on ZebraLogic): by 3.7 points (90\% interval $[2.6, 4.9]$) for OLMo's random search order, by 2.5 points ($[1.2, 3.8]$) for its randomised presentation with renamed entities, and by 1.7 points ($[-1.6, 5.1]$) for Qwen3-1.7B's random search order.
Table~\ref{tab:prespecified-b} marks these decisions.
The prompts for the confirmatory tests were fixed before generation.
Decisions are corrected with the Holm procedure~\citep{holm_simple_1979} within each model's family of pre-specified tests for each study; the post hoc shared-default comparisons of the four earlier models form a separate family.
A test therefore passes when its Holm-adjusted $p$ is at most 0.05.
The intervals we print are not adjusted, so an interval can clear its bound while the corrected test does not pass, and a two-sided test at the 5\% level corresponds to a 95\% interval, not to the 90\% intervals printed for the public benchmarks; Table~\ref{tab:benchmark-estimates} marks the four cells where this happens.
We also report descriptive summaries without multiplicity correction. These include the by-$B$ Symbolic comparisons in Figure~\ref{fig:concentration}A and the intervals shown for the random-choice instruction, higher temperature and enumerate-then-choose in Figures~\ref{fig:interventions} and~\ref{fig:appendix-interventions}. The concentration tests over all problems were pre-specified for ZebraLogic and the public benchmarks in every model that ran them, and on the Symbolic task for the two new models only.
The modal matches in Appendix~\ref{app:prospective} and the rank shares in Appendix~\ref{app:process} are post hoc summaries; both appendices give the counts.
Values that are undefined, for example breadth when a problem has no valid output, are left undefined rather than set to zero.

\paragraph{Sets of pre-specified tests.}
\label{app:studies}
Table~\ref{tab:studies} groups the pre-specified tests into six sets; each set had its analysis plan, and for the prediction tests its predictions, committed before generation.
Not every model is in every set, which is why some cells of Tables~\ref{tab:prespecified} and~\ref{tab:prespecified-b} are blank: gpt-oss-120b has no fitted procedure mixture and no specified-order runs on the controlled tasks, and the two new models ran the tests that carry the main claims, without the candidate-count design, the stronger baselines or the follow-up conversations.

\begin{table*}[t]
  \centering\footnotesize
  \setlength{\tabcolsep}{4pt}
  \begin{tabular}{@{}r>{\raggedright\arraybackslash}p{3.0cm}>{\raggedright\arraybackslash}p{3.1cm}>{\raggedright\arraybackslash}p{6.6cm}>{\raggedright\arraybackslash}p{2.0cm}@{}}
    \toprule
     & Set & Models & Pre-specified tests & Validity rule \\
    \midrule
    1 & Cueing and history & three main models & cued success by rank; history & two-sided \\
    2 & Procedures and interventions & three main models & held-out prediction; reversed orders; randomised presentation on the Symbolic task, with renamed entities and against random choice; temperature $+0.5$; the original ZebraLogic runs & two-sided \\
    3 & Stated orders, scale and training stages & three main models, gpt-oss-120b, Olmo~3 7B Think & order factorial; specified and random search order; answering without written reasoning; the gpt-oss-120b tests; training stages & two-sided \\
    4 & ZebraLogic re-run & four earlier models & concentration, randomised presentation (also against random choice), specified and random search order on ZebraLogic, with the final reader fixed beforehand & two-sided \\
    5 & Further tests & four earlier models & later ZebraLogic run (declared value order, cueing); candidates or mentions; stronger baselines; components of randomised presentation; recommended decoding; reduced reasoning; another solution; system-message control for the training stages; calendar, colouring and trip benchmarks & one-sided$^{a}$ \\
    6 & New models & Qwen3-30B-A3B, Gemma & every test marked for them in Tables~\ref{tab:prespecified}, \ref{tab:prespecified-b} and~\ref{tab:benchmark-estimates}; the nine shared-default pairs that include one of them & one-sided \\
    \bottomrule
  \end{tabular}
  \caption{\textbf{Sets of pre-specified tests}, with their models and the test of a validity change or of cued success that each set specified before generation.
  $^{a}$The two-sided results are reported as secondary.}
  \label{tab:studies}
\end{table*}

\paragraph{Decisions that changed after results were available.}
\label{app:corrections}
Tables~\ref{tab:prespecified} and~\ref{tab:prespecified-b} list every pre-specified test on the controlled tasks whose result the paper reports, with the criterion it had to meet, its outcome for each model and the section that discusses it; Table~\ref{tab:benchmark-estimates} does the same for the public benchmarks, and every other analysis in the paper is descriptive.
Sixteen pre-specified decisions differ from the first analysis of their data, all towards support.
Four are the validity decisions above; they can only move in this direction, because the one-sided rule asks less than the two-sided test.
Twelve follow corrections of analysis code that had not tested the stated criterion.
The colouring analysis compared two different encodings of a colouring and returned exactly zero for every model (two decisions: corrected, the two gpt-oss models reject, and Qwen3-1.7B and OLMo still do not).
Cued success and validity were tested against zero instead of the stated margin (six Symbolic decisions of the two new models, whose $B=2$ problems were also resampled singly instead of by structural class, and gpt-oss-20b's cued-success decision on trip planning).
Qwen3-30B-A3B's ZebraLogic answers were compared with predictions and targets stored in a different encoding, which gave shares of exactly zero (three decisions).
Of the 75 corrections and deviations recorded for sets 5 and 6, 47 changed no decision.

\section{Search procedures}
\label{app:procedures}

\begin{figure}[t]
  \centering
  \includegraphics[width=\columnwidth]{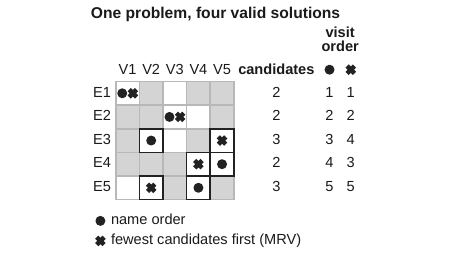}
  \caption{\textbf{One Symbolic problem, two search orders.}
  Each entity (row) takes a different value (column); grey cells are excluded, and four solutions remain.
  Name order and fewest candidates first visit the entities in different orders (right) and reach different first solutions, outlined where they differ.}
  \label{fig:worked-example}
\end{figure}

Seven procedures choose an entity and then try its remaining values in the declared order, in a depth-first search that returns the first complete solution.
They differ in which entity they fill next:
\begin{itemize}
  \item \textbf{name order}, $E_1$ to $E_5$;
  \item \textbf{listing order}, the order in which entities' facts are listed;
  \item \textbf{static MRV}, fewest initial candidates first, with ties broken by name or by listing order;
  \item \textbf{dynamic MRV}, fewest remaining candidates given the assignments so far, with ties broken by name or by listing order;
  \item \textbf{query first}, the entity whose value the prompt asks for (Appendix~\ref{app:tasks}) and then name order.
\end{itemize}
The eighth, a value-major procedure, instead takes values in declared order and gives each to the first entity, by name, that can take it.
Figure~\ref{fig:worked-example} shows a problem on which name order and static MRV visit the entities in different orders and reach different first solutions.

Each model's primary procedure was selected on earlier data and fixed before the prospective test.
It is static MRV with ties broken by listing order for Qwen3-1.7B, and static MRV with ties broken by name for OLMo, gpt-oss-20b, gpt-oss-120b and the two new models.
The new models selected it on their own free generation for the 72 main problems, in the unmodified and the relabelled presentation, and committed their 192 predictions before any held-out response was generated.

\paragraph{Prospective test.}
\label{app:prospective}
The 48 problems were drawn by a fixed rule that took structural classes and their instances in hash order and kept, for each class, the first instance not used in any earlier run for which all four presentations exist (24 classes at $B=4$, all those with such an instance, and the first 24 at $B=6$).
Twenty-eight of these classes also occur among the 72 main problems (22 at $B=4$, 6 at $B=6$), so the test is held out at the level of problem instances, and at the level of classes for the other twenty.
The overlap at $B=4$ could not be avoided: the census has only 27 classes with four solutions and the main set uses 24 of them, whereas 18 of the 24 six-solution instances come from new classes.
Each instance is shown as generated (facts in name order) and with two value relabellings, one that changes the MRV prediction but not name order's and one the reverse; the fourth presentation reverses the listing order of all 48 instances, and also relabels the values of 11 of them, so that breaking ties by name and by listing order give different predictions.
We fixed all eight procedures' predictions for the 192 presentations before generation.
The test uses the presentations on which a model's primary procedure and name order predict different solutions; because the primary procedures differ in tie-breaking, this gives 154 presentations for Qwen3-1.7B and 149 for each of the other five models.
gpt-oss-120b's primary procedure was selected on its own earlier free-generation data, before its prospective outcomes were read.
On the discriminating presentations, 48\%, 48\% and 42\% of outputs match the MRV-like and 11\%, 12\% and 16\% the name-order prediction (Qwen3-1.7B, OLMo, gpt-oss-20b); over all 192, 49\%, 49\% and 44\% match the MRV-like and 19\%, 22\% and 23\% the name-order prediction.
Where static and dynamic MRV predict different solutions, Qwen3-1.7B's outputs matched the static prediction more often (by 23 points), whereas OLMo (8 points) and gpt-oss-20b (1 point) showed no resolved difference (Table~\ref{tab:secondary}), and breaking ties by listing order mattered only for Qwen3-1.7B.
In a post hoc, descriptive check of all six models, we asked whether each prediction matched the most frequent valid answer among the 32 requests of a presentation; a predicted solution earned $1/t$ credit when it was one of $t$ tied modes, and zero otherwise.
Out of 192 presentations the MRV-like prediction earned 124, 132, 121, 122, 132 and 127 (63.0 to 68.8\%) for Qwen3-1.7B, OLMo, gpt-oss-20b, gpt-oss-120b, Qwen3-30B-A3B and Gemma, and name order earned 37, 42.5, 48, 36, 37 and 67.5 (18.8 to 35.2\%).
Because 96 presentations had four solutions and 96 had six, a uniform guess would match on average in $(1/4+1/6)/2\approx20.8\%$ of presentations.
The MRV-like match rate was 69 to 90\% on the presentations as generated and 31 to 55\% on those designed to distinguish tie rules.
The table below gives each share with its 95\% interval, from 2,000 resamples of the 48 problems that keep each problem's four presentations together, and the credit on the presentations where the two predictions differ (154 for Qwen3-1.7B, 149 for the other models).
\begin{center}\scriptsize
  \setlength{\tabcolsep}{3pt}
  \begin{tabular}{@{}llrcr@{}}
\toprule
 & & \multicolumn{2}{c}{All 192 presentations} & Differing \\
\cmidrule(lr){3-4}
Model & Prediction & Credit & Share, \% & Credit \\
\midrule
Qwen3-1.7B & MRV-like & 124 & 64.6 [57.6, 71.4] & 96/154 \\
 & name order & 37 & 19.3 [14.3, 24.7] & 9/154 \\
OLMo & MRV-like & 132 & 68.8 [62.5, 75.0] & 99/149 \\
 & name order & 42.5 & 22.1 [16.1, 28.6] & 9.5/149 \\
gpt-oss-20b & MRV-like & 121 & 63.0 [55.7, 70.6] & 90/149 \\
 & name order & 48 & 25.0 [18.5, 32.0] & 17/149 \\
gpt-oss-120b & MRV-like & 122 & 63.5 [56.3, 70.3] & 95/149 \\
 & name order & 36 & 18.8 [13.8, 24.0] & 9/149 \\
Qwen3-30B-A3B & MRV-like & 132 & 68.8 [61.7, 75.5] & 106.5/149 \\
 & name order & 37 & 19.3 [14.1, 24.5] & 11.5/149 \\
Gemma & MRV-like & 127 & 66.1 [58.9, 72.9] & 89/149 \\
 & name order & 67.5 & 35.2 [28.1, 42.7] & 29.5/149 \\
\bottomrule
\end{tabular}
\end{center}
When the main models are asked instead to list every solution, the first solution they list is the MRV-first solution in 70 to 80\% of the lists with at least two valid solutions (205 to 234 lists per model) and the name-order-first solution in 57 to 64\%.

\paragraph{Held-out structural classes.}
Twenty of the 48 prospective instances come from structural classes that occur in no earlier run.
Restricted post hoc to them, the share of all requests on the MRV-like prediction exceeds that on the name-order prediction by 38.0, 37.7, 25.3 and 37.4 points for Qwen3-1.7B, OLMo, gpt-oss-20b and gpt-oss-120b, with every interval above zero when the 20 classes are resampled (Table~\ref{tab:secondary}); this check has no multiplicity family.

\paragraph{Candidate count versus mention count.}
\label{app:dc}
In the main problems every exclusion fact both mentions an entity and removes one of its candidates, so fewest-candidates-first and most-mentioned-first are one rule.
A further design separates them on 42 instances, one from each of 42 structural classes that occur neither among the main problems nor in the prospective test, each shown in four presentations with 32 requests per presentation and model.
In R1, some exclusions of an entity that does not have the fewest candidates are restated in a second wording (``V3 is not assigned to E2''), which adds mentions without changing any candidate set; in D1, one entity's exclusions are replaced by a single disjunctive fact (``E4 is V1 or V5''), which gives it both the fewest candidates and the fewest mentions.
In both, static MRV and a mention-first procedure, each with the model's own tie rule, predict different first solutions; two control presentations apply the same changes where the predictions coincide.
The pre-specified two-sided tests compare the shares of all requests on the two predictions (Holm over the two tests per model); positive values favour candidate count, and a salience explanation predicts negative ones.
In R1 the difference is 47.4, 55.7, 32.6 and 44.6 points for Qwen3-1.7B, OLMo, gpt-oss-20b and gpt-oss-120b, and every test passes; in D1 it is 16.5 and 37.3 points for the two gpt-oss models, whose tests pass, but 3.9 points ([$-6.4$, 13.9]) for Qwen3-1.7B and $-9.2$ points ([$-20.5$, 2.3]) for OLMo, both unresolved (Figure~\ref{fig:search}A; Table~\ref{tab:secondary}).
On a subset of 35 of the 42 classes, OLMo's D1 estimate is negative with an interval excluding zero, so in that presentation OLMo may follow mentions rather than candidates.
An independent audit estimates reader error rates of 1.0 to 7.0\% across OLMo's cells and 3.0 to 3.6\% across Qwen3-1.7B's, against at most 0.3\% for the gpt-oss models, which further qualifies the two unresolved contrasts.
The two new models were not run in this design.

\paragraph{Procedure mixture.}
For a presentation $x$ of a problem with solution set $S$, procedure $k$ reaches a first solution $s_k(x)\in S$.
The mixture predicts solution $w$ with probability
\begin{equation*}
  p(w\mid x)=\sum_{k=1}^{8}\pi_k\,\mathbf 1[s_k(x)=w]+\pi_U/|S|,
\end{equation*}
with weights $\pi_k,\pi_U\ge0$ summing to one and shared across problems.
For each model, the weights were fitted by expectation--maximisation to maximise the likelihood of the valid free-generation answers to the 72 main problems in four presentations (canonical, relabelled values, reversed listing, both), with up to 48 requests per presentation for Qwen3-1.7B and 32 for the other models.
In leave-one-cluster-out evaluation on those data (clusters as in Appendix~\ref{app:stats}), the mixture's log-likelihood per valid answer is $-1.09$, $-0.92$ and $-1.11$ for the three main models, against $-1.28$, $-1.22$ and $-1.26$ for name order plus uniform choice and $-1.29$ for uniform choice alone.
On the new problems of the prospective test, with the weights frozen, it is $-1.36$, $-1.20$ and $-1.40$, against $-1.62$, $-1.65$ and $-1.61$ for name order plus uniform choice.
The weights are frozen after fitting.
To predict a new condition, each procedure is rerun on the presentation actually shown, so relabelled values, a different declared or listing order change $s_k(x)$ and hence $p(w\mid x)$ without refitting; answers are mapped back to the canonical solutions.
For randomised presentation, the forecast averages $p(w\mid x)$ over the random presentations, by exact enumeration on the Symbolic task and over 4,000 sampled presentations per variant on ZebraLogic, and the forecast breadth is the expected normalised effective number of 32 valid draws from that distribution.
For the random search order, the forecast instead assumes that the model follows the specified order, enumerating all $120\times120$ orders.
The ZebraLogic forecast uses ZebraLogic analogues of the procedures and a mixture fitted in the same way on the free-generation ZebraLogic answers collected before the intervention runs.
The weights describe the distribution of reported solutions; they do not measure how often a model runs each procedure.

\noindent Table~\ref{tab:mixture} gives the fitted weights on the Symbolic task.
For the two new models, the same mixture is fitted to their answers in the canonical and relabelled presentations only; in these, facts are listed in name order, so each pair of procedures that differ only in using names or the listing shares its weight equally.
gpt-oss-120b has no fitted mixture.
A model's primary procedure is the single procedure that matches the largest share of its valid answers when each procedure is scored on its own, whereas the weights are fitted jointly and procedures that often predict the same solution share their weight.
The largest weight therefore need not fall on the primary procedure: for OLMo and gpt-oss-20b, static MRV with ties by listing has a larger weight than their primary procedure, static MRV with ties by name.

\begin{table}[t]
  \centering\footnotesize
  \setlength{\tabcolsep}{2.5pt}
  \begin{tabular}{@{}lccccc@{}}
    \toprule
    Component & \rh{Qwen3-1.7B} & \rh{OLMo} & \rh{gpt-oss-20b} & \rh{Qwen3-30B-A3B} & \rh{Gemma} \\
    \midrule
    Name order & 0.00 & 0.00 & 0.00 & 0.00 & 0.07 \\
    Listing order & 0.03 & 0.02 & 0.00 & 0.00 & 0.07 \\
    Static MRV, ties by name & 0.06 & 0.19 & 0.16 & 0.24 & 0.18 \\
    Static MRV, ties by listing & 0.27 & 0.27 & 0.17 & 0.24 & 0.18 \\
    Dynamic MRV, ties by name & 0.11 & 0.17 & 0.10 & 0.00 & 0.15 \\
    Dynamic MRV, ties by listing & 0.04 & 0.03 & 0.02 & 0.00 & 0.15 \\
    Query first & 0.01 & 0.04 & 0.05 & 0.05 & 0.01 \\
    Value-major & 0.00 & 0.00 & 0.02 & 0.04 & 0.00 \\
    Uniform choice & 0.47 & 0.28 & 0.48 & 0.43 & 0.20 \\
    \midrule
    Procedures in total & 0.53 & 0.72 & 0.52 & 0.57 & 0.80 \\
    \bottomrule
  \end{tabular}
  \caption{\textbf{Fitted weights of the procedure mixture} on the Symbolic task.
  The main models' fits use four presentations of the 72 problems, the two new models' fits two (see text).}
  \label{tab:mixture}
\end{table}

\section{Additional results}
\label{app:additional}

\begin{figure*}[t]
  \centering
  \includegraphics[width=\textwidth]{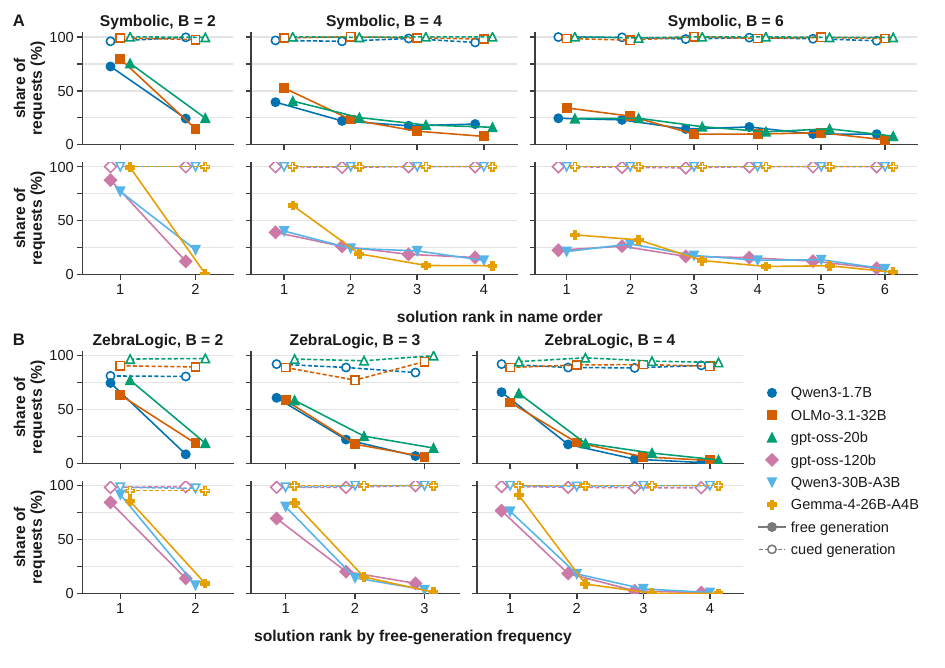}
  \caption{\textbf{Free and cued generation for every $B$.}
  Symbolic (A; $B=2,4,6$) and ZebraLogic (B; $B=2,3,4$), with the main models in the upper row of each block and the other three below.
  Share of requests giving each solution in free generation (solid, filled) and in cued generation towards it (dashed, open), by solution rank (Symbolic: name order; ZebraLogic: free-generation frequency), over all requests including invalid ones.
  Profiles are descriptive (no intervals).}
  \label{fig:appendix-rank}
\end{figure*}

\begin{table*}[t]
  \centering\footnotesize
  \begin{tabular}{@{}lrrlrrl@{}}
\toprule
 & \multicolumn{3}{c}{Symbolic (72 problems)} & \multicolumn{3}{c}{ZebraLogic (48 puzzles)} \\
\cmidrule(lr){2-4}\cmidrule(l){5-7}
Pair & Same & Chance & Excess [95\% CI] & Same & Chance & Excess [95\% CI] \\
\midrule
\multicolumn{7}{@{}l}{\textit{Pre-specified pairs (Holm family of nine)}} \\
Qwen3-1.7B $\times$ Qwen3-30B-A3B & 63.2 & 30.4 & 32.8 [23.5, 42.5] & 80.2 & 38.9 & 41.4 [30.4, 52.0] \\
OLMo $\times$ Qwen3-30B-A3B & 72.2 & 30.2 & 42.0 [32.4, 52.4] & 67.4 & 38.7 & 28.6 [14.6, 41.6] \\
gpt-oss-20b $\times$ Qwen3-30B-A3B & 75.7 & 30.6 & 45.1 [35.9, 56.9] & 80.2 & 38.9 & 41.3 [30.0, 51.2] \\
gpt-oss-120b $\times$ Qwen3-30B-A3B & 75.0 & 30.6 & 44.4 [35.2, 55.4] & 79.2 & 38.8 & 40.4 [28.5, 50.8] \\
Qwen3-1.7B $\times$ Gemma & 72.2 & 30.5 & 41.7 [31.8, 50.7] & 76.0 & 38.9 & 37.1 [24.6, 48.2] \\
OLMo $\times$ Gemma & 82.6 & 30.3 & 52.3 [43.9, 61.2] & 79.9 & 38.7 & 41.1 [28.7, 52.9] \\
gpt-oss-20b $\times$ Gemma & 73.6 & 30.5 & 43.2 [32.7, 52.6] & 92.7 & 38.9 & 53.8 [45.8, 60.3] \\
gpt-oss-120b $\times$ Gemma & 73.6 & 30.3 & 43.4 [32.3, 52.3] & 89.6 & 38.6 & 50.9 [42.4, 58.3] \\
Qwen3-30B-A3B $\times$ Gemma & 68.8 & 30.7 & 38.1 [28.7, 48.6] & 79.2 & 39.3 & 39.9 [28.6, 50.4] \\
\multicolumn{7}{@{}l}{\textit{Post hoc pairs (Holm family of six)}} \\
Qwen3-1.7B $\times$ OLMo & 75.7 & 30.5 & 45.2 [35.5, 54.5] & 66.3 & 39.3 & 27.0 [12.8, 40.9] \\
Qwen3-1.7B $\times$ gpt-oss-20b & 77.1 & 30.4 & 46.7 [37.6, 56.7] & 72.9 & 38.8 & 34.1 [22.6, 45.8] \\
Qwen3-1.7B $\times$ gpt-oss-120b & 76.4 & 30.6 & 45.8 [37.1, 55.0] & 71.9 & 39.1 & 32.8 [21.3, 44.1] \\
OLMo $\times$ gpt-oss-20b & 75.7 & 30.7 & 45.0 [35.1, 54.5] & 76.7 & 38.8 & 37.9 [25.2, 49.3] \\
OLMo $\times$ gpt-oss-120b & 79.2 & 30.3 & 48.8 [39.2, 57.7] & 79.9 & 38.9 & 41.0 [29.1, 52.1] \\
gpt-oss-20b $\times$ gpt-oss-120b & 91.0 & 30.7 & 60.3 [52.3, 70.2] & 94.8 & 39.2 & 55.6 [49.5, 61.0] \\
\bottomrule
\end{tabular}
  \caption{\textbf{Shared defaults for all 15 model pairs.}
  Same: problems on which the two models' defaults coincide (\%); chance: the same share under uniform relabelling of one model's answers (\%); excess: their difference in points, with 95\% intervals.
  The six pairs among the four earlier models are post hoc and form their own Holm family.}
  \label{tab:shared-defaults}
\end{table*}

\subsection{Recoverability at every number of solutions}
\label{app:rank}

Figure~\ref{fig:appendix-rank} repeats Figure~\ref{fig:concentration}C for every $B$ and adds the other three models.
The pattern of Sec.~\ref{sec:recoverability} holds at each of them.
For the main models on the Symbolic task, free-generation frequency at the last rank is 15 to 25\% at $B=2$ and 4 to 10\% at $B=6$, whereas cued success stays at 95 to 100\%.
On ZebraLogic, their cued success ranges from 77\% (OLMo, second-ranked of three solutions) to 100\%, far above the free-generation frequency at every rank after the first (at most 25\%).
For gpt-oss-120b, Qwen3-30B-A3B and Gemma, cued success is 99 to 100\% at every rank on the Symbolic task and 95 to 100\% on ZebraLogic, where no solution after the first is given freely in more than 20.3\% of requests.

\subsection{Shared defaults for every pair}
\label{app:shared-defaults}

Table~\ref{tab:shared-defaults} gives the shared-default comparison of Sec.~\ref{sec:concentration-results} for all 15 model pairs.
Each model contributes its first 32 requests per problem under the unmodified prompt.
\emph{Same} is the share of problems on which the two models' most frequent valid solutions coincide, with ties counted fractionally, and \emph{chance} is that share when one model's answers are relabelled uniformly over each problem's valid solutions.
The nine pairs that include Qwen3-30B-A3B or Gemma were pre-specified; the six pairs among the earlier models were compared post hoc, in a separate Holm family, and use the same procedure.
Every excess is positive with a 95\% interval above zero; after correction within each family, every test rejects this uniform-relabelling null (Holm-adjusted $p$ 0.004 for each pre-specified pair and 0.003 for each post hoc pair).
Rejecting it shows that two models favour the same solutions more often than uniform choice would produce, not that one model's choice depends on the other's: two models that independently preferred the same ordering would also exceed it, which is the reading we propose in Sec.~\ref{sec:search}.
Symbolic intervals resample 51 structural classes (the three $B=2$ classes and 24 at each of $B=4$ and 6), ZebraLogic intervals the 48 source puzzles; the other Symbolic analyses use the cluster counts of Appendix~\ref{app:stats}.

\subsection{Scale and training stages}
\label{app:stages}

\paragraph{Scale.}
gpt-oss-120b's Symbolic outputs are concentrated (modal share 66 to 88\%), and it matches its own predicted procedure more often than gpt-oss-20b does (by 9 points; Table~\ref{tab:secondary}); Qwen3-30B-A3B concentrates and follows the MRV-like rule like Qwen3-1.7B (Secs.~\ref{sec:concentration-results} and~\ref{sec:predict}).
These comparisons show persistence at the sizes we tested; size is confounded with architecture and training, so they do not show a scaling trend.

\paragraph{Training stages.}
Table~\ref{tab:stages} gives the three Olmo~3 7B Think checkpoints of Sec.~\ref{sec:robustness} on the 72 Symbolic problems.
The SFT checkpoint already concentrates its answers (modal share 64\%) and succeeds almost always when cued (99.9\%); in the same lineage, \citet{karouzos_where_2026} find that the Think line loses most of its semantic diversity at SFT.
The checkpoint after preference tuning (DPO) has a higher modal share than SFT (by 6 points) and a larger MRV-like contrast (by 8 points), but it also uses a different default system prompt, so this comparison changes checkpoint and prompt together and does not isolate a contribution of preference tuning.
After RLVR, neither quantity shows a statistically resolved change ($-0.1$ and $-3$ points, intervals including zero; Table~\ref{tab:secondary} gives the intervals of all stage contrasts).
We also crossed checkpoint and system prompt, running the SFT and DPO checkpoints each with both prompts.
With the prompt held fixed, the checkpoint contrast averaged over the two prompts is $+1.7$ points for modal share (95\% interval $[-0.8, 4.2]$) and $+2.0$ points for the MRV-like contrast ($[-2.2, 6.3]$), and neither is resolved after Holm correction (adjusted $p=0.43$).
The prompt itself matters: averaged over the two checkpoints, the DPO system prompt raises modal share by 4.22 points and the MRV-like contrast by 6.25 points (secondary comparisons, unadjusted, with intervals above zero), a reminder that prompt format alone can change how concentrated the answers are~\citep{yun_price_2025}.
The native difference therefore cannot be attributed to the weights.
In all, eight tests on the native checkpoints were pre-specified as one family: from SFT to DPO, modal share and the MRV-like contrast change and value-label following and cued success do not, and from DPO to RLVR none of the four changes; neither of the two tests with the prompt held fixed is supported.
The native MRV-like difference is also sensitive to the answer reader: with the error rates of an independent audit, its interval could include zero.
Three further checkpoints of the same lineage, Olmo~3 7B Base and the RL-Zero Mix and Math checkpoints, could not be evaluated: each failed both pre-specified admission thresholds, with 7 to 17\% valid outputs and 23 to 67\% of responses still truncated after the retry.

\begin{table}[t]
  \centering
  \resizebox{\columnwidth}{!}{\begin{tabular}{lccc}
\toprule
Stage & Modal share (\%) & Breadth (\%) & Cued success (\%) \\
\midrule
SFT & 64.4 [60.4, 68.2] & 58.7 [54.2, 63.2] & 99.9 [99.7, 100.0] \\
DPO & 70.2 [65.9, 74.5] & 53.4 [48.8, 58.5] & 99.3 [98.7, 99.8] \\
RLVR & 70.1 [66.0, 74.3] & 53.4 [48.7, 58.2] & 99.7 [99.4, 100.0] \\
\bottomrule
\end{tabular}}
  \caption{\textbf{Olmo~3 7B Think across training stages.}
  Modal share, breadth (Eq.~\ref{eq:breadth}) and cued success in percent, with 95\% intervals, on 72 Symbolic problems per stage (cued success: 288 targets, 4 requests each).
  No base checkpoint was evaluated.
  The SFT and DPO checkpoints also differ in their default system prompt.}
  \label{tab:stages}
\end{table}

\subsection{History}
\label{app:history}

For OLMo, we kept an earlier answer in the conversation and then sampled again, on the 24 problems at each of $B=2$ and 4.
The earlier answer was either the model's greedy answer or a sampled alternative, the first valid answer among 16 sampled ones that differed from the greedy answer.
Of the 48 problems, two had no valid greedy answer, and for three more none of the 16 sampled answers was a different valid solution, so the comparisons use 46 and 43 problems.
Compared with a fresh conversation, the free-generation share of the earlier answer rose by 15.1 points for the greedy answer and by 33.0 points for the sampled alternative, with intervals above zero (Table~\ref{tab:secondary}).
Cued success for the remaining solutions changed by $+0.9$ points (90\% interval $[-0.1, 1.9]$) and $-0.3$ points ($[-1.2, 0.5]$), within the pre-specified $\pm5$-point margin.

\subsection{Answering without written reasoning}
\label{app:answer-only}

In this condition the prompt asks for the final answer line only, any reasoning channel is closed, and the answer line is prefilled with \texttt{FINAL: E1=} in every model, so the first entity of the answer is supplied; the model then has 48 new tokens and no retry.
Validity fell by 86 points (Qwen3-1.7B), 57 (OLMo), 57 (gpt-oss-20b) and 74 (gpt-oss-120b) compared with free generation.
The MRV-like contrast dropped by 49, 50 and 98 points for OLMo, gpt-oss-20b and gpt-oss-120b, with all intervals above zero.
For Qwen3-1.7B it could not be estimated reliably, because only six problems had enough valid answers.
This condition changes the prompt, the token budget and the start of the answer together; because the prefilled line begins with $E_1$, the answer must also be written in name order, which may itself favour the name-order solution.
We also reduced the reasoning within each model's normal format: thinking disabled for Qwen3-1.7B and low reasoning effort for gpt-oss-20b and gpt-oss-120b.
All six pre-specified tests pass: the default's excess over the uniform expectation is 23.3, 23.8 and 41.3 points, and on the held-out presentations the MRV-like minus name-order share is 8.7, 29.5 and 44.0 points (Table~\ref{tab:secondary}).

\subsection{Process evidence}
\label{app:process}

These three exploratory analyses fall outside the pre-specified family of tests and its multiplicity correction.
They describe the written reasoning and do not establish how the models compute their answers.
Reasoning is measured in native tokens over Qwen3-1.7B's thinking, gpt-oss-20b's analysis channel and OLMo's response up to the \texttt{FINAL} line.

\paragraph{Commitment along the reasoning.}
For each of 24 problems per model (12 at each of $B=4$ and 6), we took the first valid free-generation response that ended normally, cut its reasoning at 10\%, 20\%, \dots, 90\% of its length, and sampled 16 continuations from each cut.
The \emph{commitment} at a cut is the share of continuations whose final answer is the source response's solution.
The median commitment first reaches 90\% at 50\% ($B=4$) and 70\% ($B=6$) of the reasoning for OLMo, at 80\% and 80\% for gpt-oss-20b, and at 40\% and 70\% for Qwen3-1.7B.
We also annotated the first point at which a response introduces an assignment as a trial rather than a deduction (for example, ``Suppose E4 = V2''), with model assistance and an author checking 20\% of the annotations.
Between the last cut before this first trial and the first cut after it, commitment rises by 18 points for OLMo (23 responses), 11 for gpt-oss-20b (24) and 44 for Qwen3-1.7B (24), each with an interval above zero (Table~\ref{tab:secondary}).
The first trial assignment is the one the primary procedure would try first in 18 of 23 OLMo responses, 10 of 24 gpt-oss-20b responses and 13 of 24 Qwen3-1.7B responses.
Commitment to the final solution therefore builds up over much of the reasoning, with a step at the first trial assignment, rather than being fixed at the start.

\paragraph{Reasoning length and search effort.}
For each presentation, we counted the dead ends (failed partial assignments) that the model's primary procedure visits before its first solution, and the same count for name order.
On valid answers that match the primary procedure's prediction (31,589 responses from the main free-generation data in four presentations, the prospective test and the value-label data of Sec.~\ref{sec:declaration}), log reasoning length rises with the primary procedure's dead-end count, adjusting for $B$, by 0.023 per dead end for OLMo and 0.024 for gpt-oss-20b, with intervals above zero; for Qwen3-1.7B the interval includes zero (0.013, [$-0.004$, 0.027]).
Name order's dead-end count shows no association in any model.
Predictions of length from the two counts have almost the same error, and in pairs of presentations that differ only in value labels the association with the primary procedure's count is resolved in only one of six model and pair-type combinations (OLMo, 0.057; Table~\ref{tab:secondary}).
This is therefore weak evidence that the length of the written reasoning follows the primary procedure's search effort.

\paragraph{Answers outside the predictions.}
We pooled the ranks of 48,421 valid answers from the four free-generation presentations, the prospective test and the value-order experiment. For each ordering, we counted how often an answer outside the first rank fell at rank~2. The table gives the rank-2 count divided by the number of answers outside rank~1, separately for each ordering. This post-hoc summary is descriptive (no interval).
\begin{center}\footnotesize
  \begin{tabular}{@{}lrrr@{}}
    \toprule
    Model & $B$ & MRV-like & Name order \\
    \midrule
    Qwen3-1.7B & 4 & 1,630/3,604 & 1,921/5,087 \\
               & 6 & 1,532/3,804 & 1,397/5,519 \\
    OLMo       & 4 & 1,147/2,426 & 1,425/3,436 \\
               & 6 & 814/2,452   & 1,292/4,062 \\
    gpt-oss-20b & 4 & 1,251/3,054 & 1,554/3,892 \\
                & 6 & 965/3,338   & 1,169/4,554 \\
    \bottomrule
  \end{tabular}
\end{center}
Under uniform sampling over the remaining $B-1$ solutions, the rank-2 share would be 33.3\% at $B=4$ and 20.0\% at $B=6$. All three models exceeded these references under both orderings. The MRV-like ordering had the higher rank-2 share in every comparison, with the largest difference for Qwen3-1.7B at $B=6$ (40.3\% versus 25.3\%).
We also fitted a noisy version of the primary procedure, which with probability $\delta$ replaces the next entity by a random unassigned one and with probability $\varepsilon$ replaces the next value by a random remaining consistent value; its first-solution distribution is computed exactly and fitted by maximum likelihood to the data used for the procedure mixture.
In leave-one-cluster-out evaluation, its log-likelihood per valid answer is $-1.07$, $-1.00$ and $-1.15$ for Qwen3-1.7B, OLMo and gpt-oss-20b, against $-1.09$, $-0.92$ and $-1.11$ for the mixture (Appendix~\ref{app:procedures}); on the prospective-test problems it is $-1.28$, $-1.29$ and $-1.44$, against $-1.36$, $-1.20$ and $-1.40$.
Noisy search had the higher held-out log-likelihood for Qwen3-1.7B, and the mixture for OLMo and gpt-oss-20b; a ranking of fits alone does not show which mechanism produced the answers outside the predictions.
Because noisy search is a noisy version of the same ordered search, its better fit for Qwen3-1.7B does not count against an ordered-search explanation.

\subsection{Larger temperature increase}
\label{app:temperature}

Raising the temperature by 0.5 instead of 0.25 reduced validity by 87 points for OLMo and 91 points for gpt-oss-20b and did not change it for Qwen3-1.7B.
Qwen3-1.7B's sampling truncates the tail of the distribution (top-$k$ 20, top-$p$ 0.95), whereas OLMo and gpt-oss-20b sample from the full distribution (Table~\ref{tab:decoding}), which may explain the difference.
gpt-oss-20b starts from its recommended setting and loses as much, so the collapse does not come from sampling OLMo away from its recommended one; the validity-preserving temperature of Appendix~\ref{app:baselines} raises the temperature with the tail truncated.
This condition covers only part of the problems, so we report it only here.

\subsection{Stronger baselines}
\label{app:baselines}

We compared randomised presentation with two stronger ways of asking for diverse answers, on the 72 Symbolic problems and the 48 ZebraLogic variants with 32 requests each.
Verbalized Sampling~\citep{zhang_verbalized_2026} adds its standard system prompt, which asks for five responses, each with a text and a probability, sampled from the full distribution; we keep one candidate per request, drawn in proportion to the stated probabilities.
The validity-preserving temperature is the higher of two raised temperatures (the model's usual temperature plus 0.5 or plus 1.0, each with min-$p$ 0.1) whose validity change passes the margin; it is undefined when neither does.
Pre-specified tests compare the breadth gain of randomised presentation over the unmodified prompt with that of each baseline (Holm over four tests per model).
On the Symbolic task, randomised presentation broadens more than Verbalized Sampling by 23.4, 24.5 and 24.1 points for Qwen3-1.7B, OLMo and gpt-oss-20b, and more than the validity-preserving temperature by 26.9, 33.9 and 27.9 points; all six tests pass (Table~\ref{tab:secondary}).
On ZebraLogic, the difference from Verbalized Sampling is 26.4 and 17.4 points for Qwen3-1.7B and OLMo, whose tests pass, but $-2.5$ ([$-7.9$, 2.7]) and $-6.9$ ([$-14.1$, 0.3]) points, unresolved, for gpt-oss-20b and gpt-oss-120b; the difference from the validity-preserving temperature is 23.9, 12.1 and 17.5 points for Qwen3-1.7B, OLMo and gpt-oss-120b, whose tests pass, and undefined for gpt-oss-20b, whose raised temperatures both lost too much validity.
gpt-oss-120b has no Symbolic comparison.
The ZebraLogic comparison with Verbalized Sampling depends on the answer reader: an audit estimated that answers under Verbalized Sampling are misread 2.8 to 19.5 points more often than answers under randomised presentation.
The two models whose tests pass are those with the smaller differences in misreading (2.8 points for Qwen3-1.7B and 5.9 for OLMo, against 18.6 and 19.5 for gpt-oss-20b and gpt-oss-120b).
We did not run methods that select a diverse set from many candidates for one input, such as diverse Minimum Bayes Risk decoding~\citep{jinnai_generating_2024}: they need all candidates in one place and can return only solutions that sampling has already reached, whereas here each request is answered on its own.

\subsection{Components of randomised presentation}
\label{app:pa}

Randomised presentation relabels the values, reorders the facts and, in its variant, renames the entities.
We randomised each of these alone on the 72 main problems (32 requests each), with the forecasts of the procedure mixture fixed before generation.
Randomising the value labels alone raises breadth by 25.4, 36.2 and 30.7 points for Qwen3-1.7B, OLMo and gpt-oss-20b, with the validity change passing its margin (Tables~\ref{tab:prespecified-b} and~\ref{tab:secondary}).
As the models' tie rules predict, randomising the entity names broadens OLMo's and gpt-oss-20b's answers more than randomising the order of the facts (by 8.3 and 9.2 points), whereas for Qwen3-1.7B, which breaks ties by listing order, the forecast opposite difference is not resolved (1.9 points, [$-2.4$, 6.3]).
An audit estimates reader error rates of 2.0 to 2.9\% in these arms for Qwen3-1.7B and OLMo.

\subsection{Asking for a different solution}
\label{app:it}

\begin{table*}[t]
  \centering\footnotesize
  \resizebox{\textwidth}{!}{\setlength{\tabcolsep}{4pt}
\begin{tabular}{@{}llrcccccc@{}}
\toprule
Testbed & Model & $B$ & \shortstack{Unmodified\\prompt} & \shortstack{Randomised\\presentation} & \shortstack{Random\\search order} & \shortstack{Random-choice\\instruction} & \shortstack{Temperature\\$+0.25$\textsuperscript{\(\dagger\)}} & \shortstack{Enumerate,\\then choose} \\
\midrule
Symbolic & Qwen3-1.7B & 2 & 75 / 98 / 98 & 98 / 99 / 100 & 95 / 97 / 100 & 75 / 98 / 100 & 77 / 96 / 100 & 62 / 97 / 92 \\
 &  & 4 & 61 / 98 / 91 & 86 / 98 / 100 & 86 / 98 / 99 & 60 / 99 / 90 & 64 / 97 / 95 & 52 / 99 / 80 \\
 &  & 6 & 41 / 98 / 70 & 77 / 97 / 99 & 73 / 98 / 97 & 37 / 99 / 64 & 41 / 99 / 74 & 34 / 100 / 59 \\
 & OLMo-3.1-32B & 2 & 65 / 93 / 98 & 98 / 91 / 100 & 97 / 96 / 100 & 69 / 88 / 98 & 73 / 89 / 100 & 78 / 76 / 100 \\
 &  & 4 & 49 / 92 / 86 & 85 / 93 / 100 & 91 / 95 / 100 & 51 / 90 / 89 & 56 / 91 / 89 & 76 / 71 / 98 \\
 &  & 6 & 34 / 91 / 67 & 78 / 92 / 99 & 85 / 96 / 100 & 35 / 91 / 73 & 41 / 90 / 75 & 66 / 67 / 91 \\
 & gpt-oss-20b & 2 & 73 / 100 / 98 & 96 / 100 / 100 & 97 / 99 / 100 & 74 / 100 / 100 & 79 / 100 / 100 & 93 / 98 / 100 \\
 &  & 4 & 61 / 100 / 95 & 87 / 100 / 100 & 89 / 99 / 100 & 62 / 100 / 96 & 68 / 100 / 96 & 81 / 98 / 98 \\
 &  & 6 & 45 / 100 / 78 & 81 / 100 / 100 & 84 / 99 / 100 & 46 / 100 / 83 & 54 / 100 / 87 & 70 / 97 / 95 \\
\midrule
ZebraLogic & Qwen3-1.7B & 2 & 63 / 86 / 79 & 89 / 78 / 100 & 76 / 84 / 100 & 62 / 86 / 75 & 61 / 82 / 79 & -- \\
 &  & 3 & 62 / 89 / 83 & 72 / 91 / 96 & 64 / 93 / 83 & 71 / 91 / 100 & 66 / 88 / 83 & -- \\
 &  & 4 & 41 / 89 / 64 & 75 / 87 / 92 & 54 / 94 / 88 & 42 / 91 / 59 & 42 / 91 / 70 & -- \\
 & OLMo-3.1-32B & 2 & 76 / 82 / 96 & 88 / 81 / 100 & 94 / 80 / 100 & 82 / 79 / 100 & 81 / 74 / 100 & -- \\
 &  & 3 & 62 / 80 / 92 & 59 / 78 / 88 & 72 / 86 / 100 & 64 / 77 / 96 & 55 / 71 / 88 & -- \\
 &  & 4 & 52 / 90 / 86 & 74 / 88 / 95 & 80 / 84 / 97 & 57 / 81 / 91 & 57 / 78 / 86 & -- \\
 & gpt-oss-20b & 2 & 76 / 96 / 100 & 91 / 96 / 100 & 95 / 98 / 100 & 73 / 95 / 98 & 84 / 88 / 100 & -- \\
 &  & 3 & 73 / 96 / 96 & 77 / 98 / 100 & 85 / 97 / 100 & 80 / 96 / 100 & 71 / 90 / 96 & -- \\
 &  & 4 & 49 / 98 / 84 & 72 / 98 / 98 & 85 / 99 / 98 & 59 / 96 / 92 & 62 / 94 / 91 & -- \\
\midrule
\multicolumn{9}{@{}l}{\(\dagger\) On ZebraLogic, the higher-temperature column uses the original runs. The other columns use the re-run.} \\
\bottomrule
\end{tabular}}
  \caption{\textbf{Absolute breadth, validity and coverage by condition.}
  Each cell gives breadth (Eq.~\ref{eq:breadth}; among valid answers) / validity (share of all requests) / Coverage@32 (distinct valid solutions among the 32 requests, as a share of $B$, which does not condition on validity), in percent, over the first 32 requests per problem. Means cover 24 Symbolic problems per $B$ and 24, 8 and 16 ZebraLogic variants at $B=2$, 3 and 4. Dashes mark a condition that was not run.}
  \label{tab:absolute}
\end{table*}

Cueing adds facts that single out a solution; a user would more simply ask for another answer.
We ran eight independent conversations per problem and model for the four earlier models: after the unmodified prompt, each conversation asked up to $\min(B,4)-1$ times for ``a different solution'' that satisfies every constraint and differs from every solution already given.
At the second turn, the conversations covered 56.0 to 60.9\% of the Symbolic solution sets and 60.4 to 75.1\% of the ZebraLogic sets, more than two requests under the unmodified prompt or under randomised presentation covered on the same cases, for every model.
A pre-specified test asked whether each new answer is the next solution of the model's ordered search, restarted without the solutions already given, compared with a uniform choice among the remaining ones; it passes in three of eight model and testbed combinations (Qwen3-1.7B on ZebraLogic, both gpt-oss models on the Symbolic task).
Asking for alternatives therefore broadens the answers, but it does not show that the models continue the same search.

\subsection{Remaining concentration on ZebraLogic}
\label{app:zebra-residual}

\begin{figure}[t]
  \centering
  \includegraphics[width=\columnwidth]{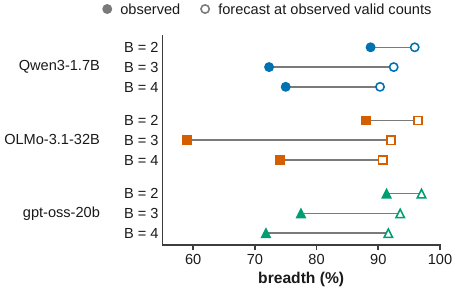}
  \caption{\textbf{Observed breadth under randomised presentation and its forecast on ZebraLogic.}
  Observed breadth among valid answers in the re-run (filled; 32 requests per variant) and the forecast fixed before the run, recomputed at the observed numbers of valid answers (open), by model and $B$.
  The forecast is not refitted; all nine 95\% intervals of the difference lie below zero.}
  \label{fig:appendix-forecast}
\end{figure}

After randomised presentation on ZebraLogic, the share of valid answers on each model's default exceeds the procedure mixture's prediction by 11.9, 13.2 and 11.2 points for Qwen3-1.7B, OLMo and gpt-oss-20b, respectively, with every interval above zero (re-run, 48 puzzles weighted equally; Table~\ref{tab:secondary}); gpt-oss-120b has no pre-specified forecast. In the original runs the excess was 14, 12 and 13 points, with all three 95\% intervals above zero.
The source puzzle's own solution does not account for it (Appendix~\ref{app:zebra-source}).
Randomised presentation on ZebraLogic leaves the house numbering, the clue wording and the value names unchanged (Appendix~\ref{app:tasks}); we did not vary these, so a preference tied to house position or to particular names remains possible.
In an exploratory split-half analysis, we added one further candidate procedure to the mixture, chosen on one half of the puzzles and evaluated on the other: the halves chose different candidates (one anchored on the clues that name a house position, the other on how often the clues mention a value), and the default's excess stayed at 3.8 to 19.2 points, with five of the six intervals above zero.
The forecast fixed before the run assumes 32 valid answers per puzzle, whereas observations use 32 requests, not all of which are valid.
The effective number computed from a finite sample is biased downwards, more so for fewer answers, so we compare observed breadth with the forecast recomputed at the observed numbers of valid answers (Figure~\ref{fig:appendix-forecast}): observed breadth remains 12.0, 15.3 and 12.1 points below the forecast (13.2, 16.2 and 12.2 points at 32 valid answers), and all nine 95\% intervals by model and $B$ are below zero, so this sampling effect explains little of the gap.
For the random search order, whose forecast is the distribution of the frozen per-request predictions, the gaps at the observed numbers are $-23.1$, $-3.7$ and $0.0$ points.
The excess share on the default is not affected by this sampling effect.

\subsection{Absolute breadth and validity}
\label{app:absolute}

Table~\ref{tab:absolute} gives absolute levels for the three main models.
On the 24 six-solution Symbolic problems, for example, randomised presentation raised Qwen3-1.7B's breadth from 41.2\% to 76.7\%, as if its valid answers had spread from about 2.5 to 4.6 of the six solutions, while validity went from 98.4\% to 97.0\%.
The Symbolic forecast fixed before the run, which assumes 32 valid draws, predicted gains on these problems of 34.3, 45.2 and 31.8 points for the three main models, against observed gains of 35.5, 44.5 and 36.1, although it put both breadth levels too high.
Averaged over the three values of $B$, the forecast gains are 27.8, 38.8 and 25.1 points, and the tested gains are 27.8, 37.4 and 30.4 points (Sec.~\ref{sec:symbolic-interventions}); with Symbolic validity at 91 to 100\% the observed numbers of valid answers are close to 32.
On ZebraLogic, the random-choice instruction itself changes breadth by $+1$ to $+5$ points (descriptive), and outputs are 2 to 6\% longer under randomised presentation (the four earlier models) and 14 to 72\% longer under a specified search order (the three main models) than under the unmodified prompt; output lengths were not recorded for Qwen3-30B-A3B and Gemma.
All Symbolic conditions use the same 72 problems. For the ZebraLogic temperature comparison, Appendix~\ref{app:zebra-original} gives the unmodified-prompt levels from the same original runs. The final reader was used throughout the ZebraLogic block.

\subsection{All six interventions}
\label{app:all-interventions}

\begin{figure*}[t]
  \centering
  \includegraphics[width=\textwidth]{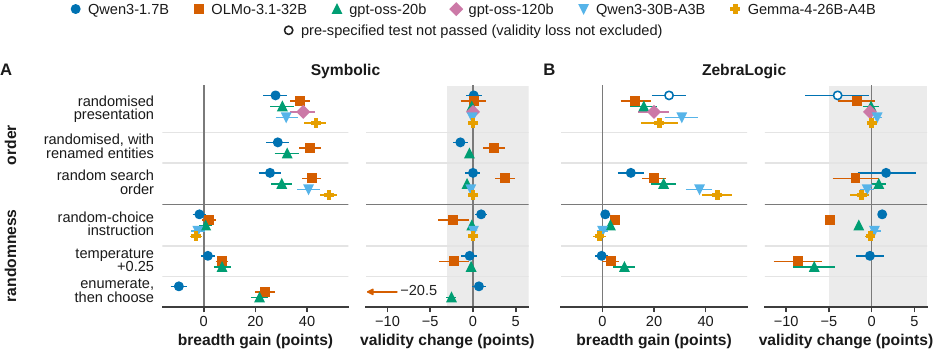}
  \caption{\textbf{Breadth and validity under all six interventions.}
  As Figure~\ref{fig:interventions}, with two further rows on the Symbolic task for the main models: randomised presentation with renamed entities, and enumerate, then choose.
  Intervals 95\% (breadth) and 90\% (validity); the arrow marks a validity change beyond the axis.}
  \label{fig:appendix-interventions}
\end{figure*}

Figure~\ref{fig:appendix-interventions} repeats Figure~\ref{fig:interventions} with the two interventions that Sec.~\ref{sec:symbolic-interventions} reports in the text only, both run on the Symbolic task with the main models.
Randomised presentation with renamed entities has a pre-specified test, which passes in each of the three; enumerate-then-choose is a descriptive comparison (Appendix~\ref{app:figure-notes}).

\section{ZebraLogic runs and answer readers}
\label{app:zebra-original}

\begin{table*}[t]
  \centering\footnotesize
  \resizebox{\textwidth}{!}{\setlength{\tabcolsep}{4pt}
\begin{tabular}{@{}llllll@{}}
\toprule
Result & Reader & Qwen3-1.7B & OLMo-3.1-32B & gpt-oss-20b & gpt-oss-120b \\
\midrule
Concentration above uniform & orig. & \textbf{33 [28, 37]} & \textbf{24 [20, 28]} & \textbf{25 [22, 29]} & 22 [17, 27] \\
 & final & \textbf{33 [28, 37]} & \textbf{24 [20, 29]} & \textbf{25 [22, 29]} & 34 [30, 38] \\
Declared-order following & orig. & \textbf{35 [27, 44]} & \textbf{30 [24, 37]} & \textbf{29 [23, 36]} & 9 [7, 11] \\
 & final & \textbf{41 [33, 51]} & \textbf{33 [25, 40]} & \textbf{36 [28, 44]} & 44 [34, 54] \\
Cued: non-default $-$ default & orig. & \textbf{$-$0.8 [$-$4.1, 2.5]} & \textbf{$-$3.9 [$-$7.5, 0.0]} & \textbf{1.2 [$-$2.3, 4.7]} & -- \\
 & final & \textbf{$-$2.0 [$-$5.2, 1.1]} & \textbf{$-$0.3 [$-$3.6, 3.0]} & \textbf{0.7 [$-$0.8, 2.4]} & -- \\
\midrule
\multicolumn{6}{@{}l}{\textit{Randomised presentation}} \\
\quad breadth gain & orig. & 25 [18, 31] & 12 [6, 18] & 17 [12, 23] & 14 [8, 21] \\
 & final & 25 [18, 31] & 12 [6, 18] & 17 [12, 22] & 23 [17, 29] \\
\quad validity change & orig. & $-$2.3 [$-$5.5, 0.7] & $-$0.1 [$-$2.4, 2.3] & $-$2.6 [$-$5.1, $-$0.1] & [$-$2.9, 2.2] \\
 & final & $-$2.5 [$-$5.7, 1.0] & $-$0.3 [$-$2.5, 2.0] & $-$0.3 [$-$1.8, 1.1] & 0.0 [$-$0.8, 0.8] \\
\quad pre-specified test met & orig. & no & \textbf{yes} & no & -- \\
 & final & no & \textbf{yes} & \textbf{yes}$^\dagger$ & -- \\
\quad gain over random-choice instruction & orig. & \textbf{22 [15, 29]} & 5 [$-$1, 13] & \textbf{12 [6, 18]} & -- \\
 & final & \textbf{24 [17, 31]} & 6 [$-$1, 13] & \textbf{13 [7, 19]} & -- \\
\midrule
\multicolumn{6}{@{}l}{\textit{Fixed specified order}} \\
\quad gain on the specified solution & orig. & $-$10 [$-$16, $-$5] & $-$4 [$-$10, 1] & 1 [$-$5, 6] & -- \\
 & final & \textbf{20 [13, 28]}$^\dagger$ & 5 [$-$1, 11] & 1 [$-$6, 7] & -- \\
\midrule
\multicolumn{6}{@{}l}{\textit{Random search order}} \\
\quad breadth gain & orig. & $-$1 [$-$7, 5] & 23 [17, 28] & 26 [21, 30] & -- \\
 & final & 14 [10, 18] & 25 [20, 31] & 24 [20, 29] & -- \\
\quad validity change & orig. & [$-$76.0, $-$62.8] & [$-$33.7, $-$25.5] & [0.8, 9.4] & -- \\
 & final & 0.1 [$-$3.2, 3.6] & 0.4 [$-$2.1, 2.8] & 0.4 [$-$0.5, 1.3] & -- \\
\quad pre-specified test met & orig. & no & no & no & -- \\
 & final & \textbf{yes}$^\dagger$ & \textbf{yes}$^\dagger$ & \textbf{yes}$^\dagger$ & -- \\
\midrule
\multicolumn{6}{@{}l}{\textit{After randomised presentation}} \\
\quad default share above forecast & orig. & 13 [7, 19] & 11 [5, 18] & 12 [7, 17] & -- \\
 & final & 14 [8, 20] & 12 [6, 19] & 13 [8, 18] & -- \\
\quad breadth $-$ forecast & orig. & $-$16 [$-$21, $-$11] & $-$18 [$-$24, $-$14] & $-$13 [$-$18, $-$9] & -- \\
 & final & $-$14 [$-$19, $-$9] & $-$18 [$-$23, $-$13] & $-$12 [$-$16, $-$8] & -- \\
\bottomrule
\end{tabular}}
  \caption{\textbf{ZebraLogic results under the original and the final answer reader.}
  All values in points, with 95\% intervals (90\% for validity changes and the cued-success difference).
  Brackets alone give the interval for an original-reader estimate.
  orig.: original reader. Bold: pre-specified test supported after correction; \dag: decision differs from the original reader. Decisions use the two-sided equivalence tests the original runs specified; under the one-sided rule, gpt-oss-20b's random search order is also supported with the original reader (Appendix~\ref{app:stats}).
  gpt-oss-120b values are secondary comparisons without a pre-specified test.}
  \label{tab:readers}
\end{table*}

\begin{table}[t]
  \centering\footnotesize
  \setlength{\tabcolsep}{3pt}
  \begin{tabular}{@{}lll@{}}
    \toprule
    ZebraLogic result & Run & Reader \\
    \midrule
    Concentration, shared defaults & re-run & final (fixed) \\
    Specified search order & re-run & final (fixed) \\
    Randomised presentation & re-run & final (fixed) \\
    Random-choice instruction & re-run & final (fixed) \\
    Random search order & re-run & final (fixed) \\
    Forecast and source checks & re-run & final (fixed) \\
    Declared value order & later run & final (fixed) \\
    Cueing (test) & later run & final (fixed) \\
    Cueing profiles (Fig.~\ref{fig:concentration}C) & original & final \\
    Higher temperature & original & final \\
    Instruction wording check & both & final \\
    New models, all conditions & own runs & final (fixed) \\
    \bottomrule
  \end{tabular}
  \caption{\textbf{Source of each ZebraLogic result.} Final (fixed): the order instruction and the final reader were fixed before generation; the original runs are read with both the pre-specified (original) reader and the final reader, chosen after their data were collected (Appendix~\ref{app:scoring}). The later run covers declared order and cueing for the four earlier models.}
  \label{tab:zebra-provenance}
\end{table}

The four earlier models have two sets of ZebraLogic runs.
The main text reports the second set (the \emph{re-run} and, for declared value order and cueing, a \emph{later run}), in which the order instruction and the answer reader were fixed before generation; only the higher-temperature comparison and the cueing profiles of Figure~\ref{fig:concentration}C come from the first set, the \emph{original runs} (Table~\ref{tab:zebra-provenance}).
This section compares the two sets and the two answer readers; the declared-order and cueing results of the original runs hold under both readers (Table~\ref{tab:readers}).

\paragraph{Order instruction in the original runs.}
The pre-specified instruction begins ``Fill the grid house by house, from House 1 to House $n$.'', and all specified-order predictions assume that the search starts from House 1.
In the original runs, every prompt with a fixed or random search order instead read ``Fill the grid house by house, from House $n$.'', where $n$ is the number of houses (3 or 4).
This wording can be read as asking the model to start from the last house; the re-run has the pre-specified wording (Appendix~\ref{app:tasks}).
Where the two readings predict different solutions, 60.7\% of gpt-oss-20b's answers to the original fixed-order prompt matched the last-house prediction, up from 47.9\% under the unmodified prompt. Its share on the first-house prediction changed from 23.0\% to 23.7\%. OLMo's share rose by 5 to 6 points for both predictions, and Qwen3-1.7B's share on the first-house prediction rose from 14.8\% to 34.9\%.
With the corrected wording, the first-house prediction matched 39\%, 59\% and 77\% of answers to the fixed-order prompt, and the last-house prediction matched 31\%, 12\% and 11\%. The gain in first-house prediction share over the unmodified prompt was 6, 32 and 53 points larger than in the original runs.
These differences combine the change of wording with run-to-run variation; for the conditions whose prompts did not change, every difference between the runs has a 95\% interval that includes zero.

\paragraph{Original-run levels.}
Under the final reader, the original runs gave breadth and validity of 56\% and 86\%, 65\% and 83\%, and 66\% and 97\% for the unmodified prompt (Qwen3-1.7B, OLMo, gpt-oss-20b), against which the higher-temperature levels in Table~\ref{tab:absolute} compare.
In these runs a higher temperature added up to 9 points of breadth but lowered validity by 8.7 points for OLMo and 6.7 points for gpt-oss-20b (Figure~\ref{fig:interventions}B).
Between the original and the final reader, breadth changes by at most 4 points except for Qwen3-1.7B under the specified orders (20 points) and gpt-oss-120b under randomised presentation (12 points), where the original reader accepted fewer than a fifth of the answers.
Validity rises by 8 to 22 points in most conditions and by 39 to 83 points for gpt-oss-120b, for the specified orders in Qwen3-1.7B and OLMo and for the higher temperature in gpt-oss-20b; a reader with the vocabulary step alone gives intermediate validity.

\paragraph{Did models favour the source puzzle's solution?}
\label{app:zebra-source}
Each public source puzzle has a unique solution, which we recovered from the full set of clues and which remains valid after clue removal.
In the re-run, read with the fixed reader, the share of valid answers giving the source solution after randomised presentation exceeds the procedure mixture's forecast by only 0.2 (95\% interval $[-5.0, 5.3]$), 0.4 ($[-5.5, 5.9]$) and 1.3 ($[-3.7, 6.5]$) points for Qwen3-1.7B, OLMo and gpt-oss-20b; gpt-oss-120b has no forecast.
Under the unmodified prompt in the original runs, rescored with the fixed reader, the source solution accounts for 32 to 37\% of the agreements between pairs of the four earlier models, below the 42\% expected if both models chose independently and uniformly among each puzzle's solutions.
Adding the source solution and a solution obtained by ordering attributes and values alphabetically to the mixture closes only 0.7 to 1.9\% of the gap between forecast and observed breadth in the re-run.
These tests measure preference for the original solution; the models' exposure to the puzzles during training remains unknown.

\subsection{Reading ZebraLogic answers}
\label{app:scoring}

The re-run uses the final reader described here, fixed before its generation.
Every ZebraLogic result in the main text, except the two that Table~\ref{tab:zebra-provenance} assigns to the original runs, therefore comes from responses generated after the reader was fixed, and a later independent audit estimated its reading error on single-answer responses at no more than 6.2\% in any model and condition (Table~\ref{tab:readers-audit}).
For each ZebraLogic response, the final answer is read from the text marked \texttt{FINAL}.
The reader first removes one attached end-of-text token and maps value spellings through a fixed vocabulary.
Complete assignments are then put in a canonical form by house, attribute and value.
Repeated statements of the same assignment count once, and two or more different assignments make the answer ambiguous.
If no complete \texttt{FINAL} assignment can be read, a fixed rule reads the answer from other layouts.
The combination of the two parts was chosen after responses of the original runs had been inspected, and the layout rule alone did not meet its pre-specified acceptance check (3 errors among 60 newly read responses, with at most 1 allowed), so we report the original runs under both readers.

\paragraph{Why the readers differ.}
The original, pre-specified reader rejected many answers only because of value spellings or layout.
About a third of the answers it marked invalid were complete answers under the fixed vocabulary.
In a post-hoc audit against 300 manually read responses, the final reader agreed with the manual reading in 282.
In 13 of the 18 disagreements it read no answer where the manual reading found one.
In 4 it read one answer where the manual reading found none or several, and in 1 it read the answer as ambiguous.
The audit samples were stratified towards responses whose reading changed, so this agreement rate is not an estimate of overall accuracy.
What the audit does establish is the direction of the remaining errors: most are missed answers, which lower measured validity.
The same reader is applied to both arms of each comparison, so a miss rate common to both arms largely cancels in a validity difference.
A miss rate that differs between arms does not cancel, and the specified-order instruction changes how answers are written.
The audit was not designed to bound such differential errors at the scale of the $\pm5$-point margin, so equivalence decisions that change between readers are conditional on the final reader.

\paragraph{Which results change.}
Table~\ref{tab:readers} compares the original and final readers on the concentration, cueing and intervention results from the original runs.
Concentration, declared-order following, cued recoverability, the breadth gains under randomised presentation, its pre-specified test for OLMo and its comparison with the random-choice instruction hold under both.
Five pre-specified decisions change, all from not supported to supported: the test for randomised presentation for gpt-oss-20b, fixed specified-order following for Qwen3-1.7B, and the test for the random search order for all three main models.
The changes for the random search order are the largest.
Under the original reader, it appeared to cut validity by tens of points for Qwen3-1.7B and OLMo (Table~\ref{tab:readers}), and Qwen3-1.7B also showed no breadth gain, on only the 31 puzzles that still had valid answers.
Under the final reader, its validity changes by less than 4 points in every model.
The specified-order instruction changes how models write their final grids, and the original reader failed to read many of those answers.
Conclusions about specified search orders on ZebraLogic therefore rest on the final reader.
Because the final reader was selected after the responses were inspected, and its layout component failed its own acceptance check, we treat the five changed decisions as sensitivity results under revised scoring, not as pre-specified confirmations; they are marked \dag{} in Table~\ref{tab:readers}.

\section{Public planning benchmarks}
\label{app:benchmarks}

\begin{table*}[t]
  \centering\scriptsize
  \setlength{\tabcolsep}{3pt}
  \begin{tabular}{@{}lrrrrrr@{}}
\toprule
Claim and task & Qwen3-1.7B & OLMo & gpt-oss-20b & gpt-oss-120b & Qwen3-30B-A3B & Gemma \\
\midrule
\multicolumn{7}{@{}l}{\textit{Concentration on a default}} \\
\quad Calendar & 18.2\,\yes & 33.5\,\yes & 40.6\,\yes & 41.3\,\yes & 34.3\,\yes & 50.8\,\yes \\
 & {\tiny [14.2, 22.4]} & {\tiny [30.4, 36.6]} & {\tiny [38.0, 43.1]} & {\tiny [37.4, 44.9]} & {\tiny [30.9, 37.6]} & {\tiny [47.3, 54.1]} \\[1pt]
\quad Colouring & 13.1\,\yes & 12.1\,\yes & 17.6\,\yes & 22.5\,\yes & 24.7\,\yes & 26.9\,\yes \\
 & {\tiny [9.3, 16.8]} & {\tiny [8.1, 16.0]} & {\tiny [14.5, 20.6]} & {\tiny [18.9, 25.9]} & {\tiny [20.5, 29.2]} & {\tiny [22.9, 31.0]} \\[1pt]
\quad Trip & -- & -- & 17.6\,\yes & 17.0\,\yes & 18.4\,\yes & -- \\
 &  &  & {\tiny [14.1, 21.0]} & {\tiny [13.3, 20.8]} & {\tiny [14.0, 22.8]} &  \\[1pt]
\addlinespace[3pt]
\multicolumn{7}{@{}l}{\textit{The named order predicts the default}} \\
\quad Calendar: earliest time & $-$22.6\,\no & 18.7\,\yes & 41.8\,\yes & 46.3\,\yes & 32.0\,\yes & 54.4\,\yes \\
 & {\tiny [$-$26.3, $-$18.8]} & {\tiny [13.5, 24.1]} & {\tiny [38.6, 44.9]} & {\tiny [41.9, 50.8]} & {\tiny [26.7, 37.4]} & {\tiny [49.3, 59.3]} \\[1pt]
\quad Colouring: fewest colours vs.\ degree & $-$1.1\,\nd & 1.0\,\nd & 17.5\,\yes & 16.5\,\yes & 5.3\,\nd & $-$12.6\,\nd$^{\S}$ \\
 & {\tiny [$-$8.5, 5.6]} & {\tiny [$-$1.6, 3.8]} & {\tiny [9.3, 25.1]} & {\tiny [6.0, 27.2]} & {\tiny [$-$6.0, 16.1]} & {\tiny [$-$24.5, $-$0.5]} \\[1pt]
\quad Trip: fewest positions first & -- & -- & 2.5\,\no & 6.4\,\no$^{\ddagger}$ & $-$6.6\,\no & -- \\
 &  &  & {\tiny [$-$3.1, 8.1]} & {\tiny [0.2, 13.0]} & {\tiny [$-$12.2, $-$0.9]} &  \\[1pt]
\quad Trip: benchmark's pair order & -- & -- & 11.8\,\yes & 9.9\,\yes & 10.0\,\yes & -- \\
 &  &  & {\tiny [7.5, 16.6]} & {\tiny [4.9, 15.4]} & {\tiny [5.9, 14.6]} &  \\[1pt]
\addlinespace[3pt]
\multicolumn{7}{@{}l}{\textit{A stated order is followed}} \\
\quad Calendar: order of times & 13.7\,\yes & 31.8\,\yes & 17.2\,\yes & 50.5\,\yes & 68.0\,\yes & 81.0\,\yes \\
 & {\tiny [10.2, 17.5]} & {\tiny [26.8, 36.4]} & {\tiny [13.8, 20.8]} & {\tiny [45.1, 55.6]} & {\tiny [60.3, 74.9]} & {\tiny [74.5, 87.3]} \\[1pt]
\quad Trip: order of cities & -- & -- & 12.4\,\yes & 18.4\,\yes & 17.1\,\yes & -- \\
 &  &  & {\tiny [7.3, 17.8]} & {\tiny [12.6, 24.8]} & {\tiny [10.6, 24.3]} &  \\[1pt]
\addlinespace[3pt]
\multicolumn{7}{@{}l}{\textit{Cued non-default solutions within the recovery margin}} \\
\quad Calendar & $-$3.6\,\yes & $-$6.9\,\no$^{\ddagger}$ & $-$9.4\,\no & $-$9.1\,\no & 0.1\,\yes & 0.1\,\yes \\
 & {\tiny [$-$8.5, 1.5]} & {\tiny [$-$9.7, $-$4.2]} & {\tiny [$-$11.9, $-$7.1]} & {\tiny [$-$12.1, $-$6.3]} & {\tiny [$-$0.2, 0.4]} & {\tiny [$-$0.1, 0.5]} \\[1pt]
\quad Colouring & $-$6.4\,\no & $-$3.1\,\yes & $-$1.3\,\yes & $-$0.5\,\yes & $-$0.7\,\yes & $-$0.3\,\yes \\
 & {\tiny [$-$10.8, $-$1.9]} & {\tiny [$-$7.8, 1.6]} & {\tiny [$-$3.0, 0.4]} & {\tiny [$-$1.3, 0.2]} & {\tiny [$-$2.6, 0.9]} & {\tiny [$-$0.8, 0.0]} \\[1pt]
\quad Trip & -- & -- & 2.6\,\yes & 0.5\,\yes & $-$8.0\,\no & -- \\
 &  &  & {\tiny [$-$0.8, 6.2]} & {\tiny [$-$3.4, 4.5]} & {\tiny [$-$14.6, $-$1.6]} &  \\[1pt]
\addlinespace[3pt]
\multicolumn{7}{@{}l}{\textit{Shuffled presentation: breadth up, validity kept}} \\
\quad Calendar & $-$5.2\,\no & 1.9\,\no & 0.6\,\no & 0.2\,\no & 0.5\,\no & 1.1\,\no \\
 & {\tiny [$-$10.3, $-$0.1]} & {\tiny [$-$0.4, 4.4]} & {\tiny [$-$1.2, 2.2]} & {\tiny [$-$1.4, 1.8]} & {\tiny [$-$1.0, 2.0]} & {\tiny [$-$0.1, 2.3]} \\[1pt]
\quad Colouring & 15.3\,\no$^{v}$ & 9.3\,\no$^{v}$ & 20.9\,\yes & 26.3\,\yes & 28.3\,\yes & 32.2\,\yes \\
 & {\tiny [10.4, 20.3]} & {\tiny [4.0, 15.0]} & {\tiny [16.7, 25.2]} & {\tiny [21.2, 31.4]} & {\tiny [23.1, 33.5]} & {\tiny [27.1, 37.4]} \\[1pt]
\quad Trip & -- & -- & 16.3\,\yes & 12.1\,\yes & 12.2\,\yes & -- \\
 &  &  & {\tiny [11.7, 20.9]} & {\tiny [7.5, 16.7]} & {\tiny [6.7, 17.7]} &  \\[1pt]
\addlinespace[3pt]
\multicolumn{7}{@{}l}{\textit{Random order: breadth up, validity kept}} \\
\quad Calendar: candidate times & 17.9\,\yes & 31.7\,\no$^{v}$ & 13.8\,\no$^{v}$ & 26.8\,\yes & 40.0\,\no$^{v}$ & 52.6\,\no$^{v}$ \\
 & {\tiny [12.9, 23.1]} & {\tiny [27.9, 35.3]} & {\tiny [11.0, 16.6]} & {\tiny [23.1, 30.4]} & {\tiny [36.9, 43.3]} & {\tiny [49.4, 55.7]} \\[1pt]
\quad Trip: cities & -- & -- & 9.1\,\yes & 7.8\,\yes & 5.2\,\no$^{\ddagger}$ & -- \\
 &  &  & {\tiny [5.8, 12.5]} & {\tiny [4.0, 11.8]} & {\tiny [0.2, 10.2]} &  \\[1pt]
\addlinespace[3pt]
\multicolumn{7}{@{}l}{\textit{Random order beats the random-choice instruction}} \\
\quad Calendar & 13.2\,\yes & 14.3\,\yes & $-$10.4\,\no & 18.2\,\yes & 36.4\,\yes & 51.6\,\yes \\
 & {\tiny [7.6, 18.5]} & {\tiny [10.8, 17.5]} & {\tiny [$-$13.8, $-$7.1]} & {\tiny [14.9, 21.4]} & {\tiny [32.9, 40.0]} & {\tiny [48.5, 54.5]} \\[1pt]
\quad Trip & -- & -- & 7.0\,\yes & 3.6\,\no & 7.6\,\yes & -- \\
 &  &  & {\tiny [4.0, 10.0]} & {\tiny [$-$0.4, 7.5]} & {\tiny [3.3, 11.9]} &  \\[1pt]
\bottomrule
\end{tabular}
  \caption{\textbf{Pre-specified tests on the three public planning tasks}, grouped by claim: estimates in points with unadjusted 90\% intervals, followed by the decision after Holm correction within each model and task (\yes passed, \no not passed, \nd two-sided test not rejected; --: not admitted or stopped at the pilot).
  Concentration: modal share minus its uniform expectation; named order and pair order: share on the predicted solution minus chance ($1/B$, or the share under a random pair order), for colouring the share on the static-MRV prediction minus that on the degree-first prediction (two-sided); stated order: change in the share on the solution the stated order reaches first; cued recovery: cued success on non-default minus default targets, which must lie above $-10$ points; shuffled presentation and random order: change in breadth, with a one-sided 5-point validity margin; last group: breadth gain of the random order minus that of the random-choice instruction.
  $^{\ddagger}$The interval clears its bound, but the adjusted $p$ exceeds 0.05. $^{\S}$Two-sided test at the 5\% level. $^{v}$Breadth rose, but the validity criterion was not met.}
  \label{tab:benchmarks}
  \label{tab:benchmark-estimates}
\end{table*}

All three benchmarks follow the design of the controlled tasks: free generation with 32 requests per instance and condition, a cued condition with eight requests per target, answers read by two independently written readers fixed before the main runs, and pre-specified tests corrected by Holm within each model and benchmark.
Validity and cued recovery use the one-sided rule of Appendix~\ref{app:stats}, fixed before generation for these studies, with margins of 5 points for validity and 10 points for cued success; two-sided equivalence versions of the same components were recorded as secondary results.
Table~\ref{tab:benchmark-estimates} lists every test with its estimate and decision.

\paragraph{Calendar scheduling.}
We used the 1,000 calendar-scheduling instances of the official NATURAL PLAN release~\citep{zheng_natural_2024} and their zero-shot prompts.
A parser reconstructs participants, working hours, half-hour busy intervals, days, duration and stated preferences, and two independent enumerators agree on every solution set; 934 instances, for which both chronological first-fit solvers return the published slot under the original constraints, were retained.
Removing only the sentences that ask for the earliest availability, and keeping every other preference as a hard constraint, leaves 193 instances with two to six valid start times, from which we drew 72 for analysis, 24 at $B=2$, 24 at $B=3$ or 4 and 24 at $B=5$ or 6, mixing single- and multi-day instances, and a disjoint pilot.
The prompt adds ``Several times may satisfy all constraints; give one.'' and the benchmark's answer format.
Conditions: the derived prompt; the participants, their busy intervals and the listed days in a random order per request (shuffled presentation); the candidate start times in a fixed order to try, chosen so that its first feasible time is not the earliest (stated order); the candidate times in a random order per request (random order); the random-choice instruction; and one cue per valid time, a sentence restricting the meeting to a window that contains only that time.
The pre-specified prediction is chronological first-fit, the earliest feasible time with days in calendar order.

\paragraph{Graph colouring.}
From the graph-colouring instances of NPHardEval~\citep{fan_nphardeval_2024} (6 to 15 vertices), we set the number of colours to each graph's chromatic number and added pre-coloured vertices, drawn from a random proper colouring, until two to six valid colourings remained; the 48 analysis graphs have two to five (7, 14, 26 and one graph with two, three, four and five).
A solution is a labelled colouring, so two colourings that differ only by an exchange of colours count as two solutions.
Two independently written solvers enumerated every solution set and had to return the same set.
All vertices share one palette, but pre-coloured neighbours remove colours from it: fewest available colours first (static MRV) visits first the uncoloured vertex with the fewest colours left after its pre-coloured neighbours are considered, whereas the most-mentioned vertex is the one with the highest degree; both rules break ties by vertex number.
The two rules predict different first colourings on 46 of the 48 graphs, and the two-sided test of static MRV against degree-first uses these.
Conditions: the derived prompt; the vertex numbers, edge order and colour letters randomised per request and mapped back (shuffled presentation); and one added pre-colouring per target colouring.

\paragraph{Trip planning.}
The NATURAL PLAN trip-planning prompts~\citep{zheng_natural_2024} leave several scored rules implicit, and with the prompts as written no model produced a valid itinerary.
Our prompts therefore state every rule the scorer enforces (the trip spans days 1 to $T$; each city is visited once, in one stay of its stated length; consecutive stays share the flight day; only the listed direct flights are used, in their stated directions; every event window is covered) and reads the itinerary from the final block of stay lines.
From the 1,600 instances, after excluding those whose reference itinerary is not the unique solution, removing one event sentence at a time in source order and keeping the first removal that leaves two to six valid itineraries, we drew 48 instances for analysis (16 each at $B=2$, $B=3$ and $B\in\{4,5,6\}$) and eight for a pilot.
The source instances visit three to ten cities, and two independently written solvers, a depth-first search and a dynamic programme over subsets of cities, enumerated the valid itineraries and had to agree.
The source writes every two-way flight pair (``X and Y'') with the earlier city of the reference itinerary first; our prompts draw each pair's order at random, and one condition keeps the source order.
Conditions: the derived prompt; the constraint sentences, the flight list and each pair's order randomised per request (shuffled presentation); a fixed order of cities to try, which is not itself a valid route (stated order); that order shuffled per request (random order); the random-choice instruction; the source pair order; and one cue per valid itinerary (``For this request, be in X on day $d$.'').
The pre-specified prediction is fewest feasible positions first: cities are taken in order of how many positions they occupy across the valid itineraries, ties broken by listing order, and each gets the earliest position consistent with a valid itinerary.
Its test compares the share of all requests on the predicted itinerary with $1/B$, so invalid answers count against the prediction; only 58\% of Qwen3-30B-A3B's answers to the derived prompt were valid, against 89\% of gpt-oss-120b's, which may explain its negative estimate.
A model entered the main study only if the two readers agreed on at least 99\% of the pilot responses either of them read, at least 30\% of its pilot responses were valid and at most 10\% remained truncated after the retry.
Qwen3-1.7B and OLMo failed this gate (0.8\% and 3.9\% valid; their invalid answers mostly missed or repeated cities), and Gemma stopped at the pilot because 12.5\% of its responses were still truncated, 11 of 16 with an unclosed reasoning section.

\paragraph{Reading Table~\ref{tab:benchmark-estimates}.}
The table prints each estimate with its unadjusted 90\% interval beside the decision after Holm correction, so three kinds of cell need a word.
In three cells ($^{\ddagger}$) the interval clears its bound but the adjusted $p$ exceeds 0.05: OLMo's cued recovery on calendars (0.056), gpt-oss-120b's fewest-positions prediction on trips (0.086) and Qwen3-30B-A3B's random city order (0.14).
Gemma's colouring contrast ($^{\S}$) is a two-sided test at the 5\% level: its 90\% interval excludes zero, but its adjusted $p$ is 0.08.
In the joint tests marked $^{v}$, breadth rose but the validity criterion was not met.
Under a random order of calendar times the validity change is $-9.5$ points (90\% interval $[-13.2, -5.9]$) for OLMo, $-4.6$ points ($[-6.6, -2.4]$) for gpt-oss-20b, $-8.6$ points ($[-11.7, -5.6]$) for Qwen3-30B-A3B and $-6.3$ points ($[-9.2, -3.7]$) for Gemma; on shuffled graphs it is $-2.2$ points ($[-5.8, 1.5]$) for Qwen3-1.7B and $-4.1$ points ($[-7.0, -1.1]$) for OLMo.
Table~\ref{tab:benchmark-validity} lists the validity change of all 24 such tests; in nine of them the lower limit lies below $-5$ points, the six cells marked $^{v}$ and the shuffled calendar prompts of Qwen3-1.7B, OLMo and gpt-oss-20b, which gave no breadth gain either.

\section{Answer readers and audits}
\label{app:readers}

\begin{table*}[t]
  \centering\footnotesize
  \setlength{\tabcolsep}{2.6pt}
  \begin{tabular}{@{}llcccccc@{}}
    \toprule
    Testbed & Conditions & Qwen3-1.7B & OLMo & gpt-oss-20b & gpt-oss-120b & Qwen3-30B-A3B & Gemma \\
    \midrule
    Symbolic & single answer & 0 to 3.0\% & 0.2 to 5.4\% & 0 to 1.2\% & 0 to 0.7\% & 0 to 0.3\% & 0\% \\
             & several candidates$^{a}$ & 1.0 to 2.6\% & 8.5 to 19.3\% & 2.1 to 3.2\% & & & \\
    Candidate count & single answer & 3.0 to 3.6\% & 1.0 to 7.0\% & 0.2 to 0.3\% & 0 to 0.1\% & & \\
    \addlinespace[2pt]
    ZebraLogic & single answer & 0.5 to 2.9\% & 2.1 to 6.2\% & 0.9 to 3.3\% & 0.4 to 1.2\% & 0 to 1.0\% & 0 to 1.0\% \\
               & Verbalized Sampling & 4.9\% & 10.1\% & 19.6\% & 19.9\% & & \\
               & follow-up turns$^{b}$ & 2.6 to 4.1\% & 3.4 to 17.8\% & 2.0 to 2.2\% & 1.8 to 3.9\% & & \\
    \addlinespace[2pt]
    Calendar & all & \multicolumn{4}{c}{two readers agree on every pilot response} & 100\%$^{c}$ & 100\%$^{c}$ \\
    Graph colouring & all & \multicolumn{4}{c}{pilot agreement only} & 100\%$^{c}$ & 100\%$^{c}$ \\
    Trip planning & all & not admitted & not admitted & 100\% (pilot) & 100\% (pilot) & 97.3 to 98.1\%$^{c}$ & pilot only \\
    \bottomrule
  \end{tabular}
  \caption{\textbf{Answer-reader audits.}
  Upper rows: estimated reading-error rates of the fixed reader from the latest independent audit, as ranges over the audited conditions.
  Lower rows: agreement between two readers.
  $^{a}$Enumerate-then-choose and Verbalized Sampling. $^{b}$Asking for a different solution (Appendix~\ref{app:it}).
  $^{c}$Agreement on the main responses that either reader read; responses that neither read are counted separately.
  Blank: not run or not audited.}
  \label{tab:readers-audit}
\end{table*}

Every reported estimate uses the reader fixed before its run; audits qualify results and do not replace that reader.
Table~\ref{tab:readers-audit} summarises the audits by testbed and model.
It separates the conditions in which a response states a single answer, which carry the main claims, from those in which it states several candidates or follows earlier turns of a conversation; the large error rates belong to the second group.
Four findings bear on specific results.
First, on ZebraLogic, answers under Verbalized Sampling are misread more often than answers under randomised presentation (by an estimated 2.8 to 19.5 points in the four earlier models), so the comparison with Verbalized Sampling (Appendix~\ref{app:baselines}) is sensitive to the reader.
Second, an independent audit of the Symbolic readings flags three contrasts: OLMo's randomised presentation with renamed entities keeps its decision under the audit's error rates, whereas OLMo's added breadth from renaming and the native DPO-minus-SFT difference in the MRV-like contrast could cross zero.
Third, for Gemma's declared-order, randomised-presentation, specified-order and random-order tests on ZebraLogic, the audited arm showed no reading error while its unmodified-prompt comparison showed two in 200 reviewed responses, and for Qwen3-30B-A3B the estimated reading error differs between randomised presentation and the random-choice instruction by more than half a point.
Fourth, the calendar and colouring readings and Qwen3-30B-A3B's trip planning rest on agreement between two readers only, with no independent transcription audit.

\section{Figure notes}
\label{app:figure-notes}

\paragraph{Figure~\ref{fig:concentration}.}
Panel A shows each model's modal share minus the modal share expected from uniform draws at the same valid sample size, one row per task and one lane per model, with the two controlled tasks on the left and the three planning tasks from public benchmarks on the right.
The Symbolic rows give this excess over the 24 problems at each $B$, with 95\% intervals; they are descriptive.
The ZebraLogic row is the pre-specified test over all 48 puzzles (95\% intervals; Table~\ref{tab:prespecified}), and the rows of the three planning tasks are their pre-specified concentration tests (Table~\ref{tab:benchmark-estimates}), whose intervals are 90\% because these tests are one-sided at the 5\% level; trip planning has the three admitted models.
Each model contributes its responses to the unmodified prompt, and panel B uses the same responses.
At Symbolic $B=2$, the intervals for the two new models resample only the three structural classes, too few clusters for a reliable bootstrap interval, so they are drawn for completeness; the four earlier models' intervals at $B=2$ resample problems.
In absolute terms, the four earlier models' defaults take 78 to 88\% of valid outputs at $B=2$, where the uniform expectation is about 57\%.
In panel B, two models share a default on a problem when their most frequent valid solutions coincide (ties counted fractionally); each cell gives the share of problems on which they do, for the Symbolic task below the diagonal and for ZebraLogic above it, with the models in the legend's order on the diagonal.
The chance level quoted in the caption relabels one model's answers uniformly over each problem's valid solutions (1,000 relabellings per problem); it is 30.2 to 30.7\% for the 15 Symbolic pairs and 38.6 to 39.3\% for the ZebraLogic pairs (Table~\ref{tab:shared-defaults}).
The nine pairs that include a new model were pre-specified; the six pairs among the four earlier models, the upper-left block of the matrix, were compared post hoc with the same procedure.
The pre-specified test of gpt-oss-120b's agreement with gpt-oss-20b (Table~\ref{tab:prespecified}) used a different baseline, which pairs one model's modal answer with another problem's (2\%); a shuffled answer need not be valid for its destination problem, so that baseline does not test independent choice within a problem and is not drawn.
Panel C shows the main models; its ZebraLogic profiles come from the original runs, and the later cueing run supplies the pre-specified comparison in Sec.~\ref{sec:recoverability}.
Symbolic solutions are ranked in name order (rank 1 is the solution a name-order search reaches first, which is not always the default), and ZebraLogic solutions by observed free-generation frequency, with ties shared, so the ZebraLogic free curve falls by construction and only its comparison with cued success is informative.
Free-generation frequency and cued success keep all requests, including invalid ones, in the denominator.
Joint movement (Sec.~\ref{sec:concentration-results}) is the share of problems on which both models' most frequent solutions differ between the unmodified presentation and the manipulated one and coincide in the manipulated one (ties counted fractionally), over the 72 problems on which both models have at least eight valid outputs in both conditions; its null relabels one model's answers by the same random permutation of the solution set in both conditions (1,000 permutations per problem).
It is a descriptive analysis of the main models.

\paragraph{Figure~\ref{fig:overview}.}
The figure shows one ZebraLogic variant, the two-solution puzzle quoted in Appendix~\ref{app:tasks}, which was chosen before its results were seen.
Its two arrangements agree on the three residents and their nationalities and differ in whether Peter (arrangement~1) or Eric (arrangement~2) loves soccer.
Each square is one executed request, 32 per model in panels A, C and D and 8 in panel B.
The runs behind each panel are those of Table~\ref{tab:zebra-provenance}.
Panel B adds the clue that leaves only arrangement~2; the clue for arrangement~1, which is not drawn, gave arrangement~1 in all 48 requests.
Of the 624 drawn requests, 20 were invalid (grey).
``Predicted'' is the pre-specified prediction of a house-by-house search that tries each value list in its stated order; it is arrangement~1 for the original value lists and arrangement~2 for the reversed ones, for all six models.

\paragraph{Figure~\ref{fig:worked-example}.}
The figure shows one problem of the Symbolic task with $B=4$, the first held-out problem on which the two procedures predict different solutions.
The grid has one row per entity and one column per value; a cell is grey when a fact excludes that value for that entity, and the count at the right of each row is the number of values left.
Name order visits the entities as they are named, and fewest candidates first visits those with two candidates before those with three; each search takes the leftmost value that leads to a complete solution, and the cells where the two first solutions differ are outlined.

\paragraph{Figure~\ref{fig:appendix-rank}.}
The three models in the lower rows use the conventions of the main models' profiles: Symbolic profiles cover 72 problems with 2,304 free and 2,304 cued requests per model, and ZebraLogic profiles 48 puzzles with 1,536 free and 1,088 cued requests, eight per target.
gpt-oss-120b's ZebraLogic profile pairs free generation in the re-run with the cued outcomes of the later run.

\paragraph{Figure~\ref{fig:search}.}
Panel A compares the shares on two predicted solutions within the same outputs; the values are differences in prediction share, not accuracy, and all shares count invalid outputs in the denominator.
Its held-out group uses 154 presentations for Qwen3-1.7B and 149 for each of the other five models.
The other two groups are the pre-specified two-sided tests of the candidate-count design (Appendix~\ref{app:dc}) for the four earlier models: on 42 further structural classes, the share on the candidate-count prediction minus that on the mention-count prediction, when some facts are restated and when one entity's exclusions are replaced by a disjunctive fact; the two new models were not run in this design.
Panel B reports, for the Symbolic task, factorial contrasts on 48 problems: with $s_c$ the share on the solution predicted for declaration order $d$ minus the share on that predicted for its reverse $d'$ in condition $c$, the declaration effect is $\tfrac14\{[s(d|L)-s(d'|L)]+[s(d|L')-s(d'|L')]\}$, where $L$ and $L'$ are the two listing orders.
This equals the average of four paired changes in share on a predicted solution when its declaration order replaces the reverse, with the listing order fixed; the listing effect is defined analogously.
The exclusion-listing order is the order of each entity's excluded values, not of whole facts.
Its ZebraLogic group is the pre-specified test of the reversed declared value order on the 48 puzzles, for all six models: the mean of two gains in share, that of the solution predicted for the original order when this order rather than its reverse is shown, and that of the solution predicted for the reversed order when the reverse is shown (the four earlier models from the later run of Table~\ref{tab:zebra-provenance}).
Panel C reports the share on the specified solution under the instruction minus that under the unmodified prompt.
The Symbolic conditions use 47 (both orders) or 48 problems and ZebraLogic uses 48 puzzles; the two new models were run with both orders and with the ZebraLogic instruction, and gpt-oss-120b was not run with specified orders on these two tasks.
The calendar and trip-planning groups are the stated-order tests of Sec.~\ref{sec:benchmarks}, with 90\% intervals: the calendar prompt lists the candidate start times in a fixed order whose first feasible time is not the earliest, and the trip prompt gives a fixed order of cities to try (Appendix~\ref{app:benchmarks}); six and three models ran them, and graph colouring has no stated-order condition.

\begin{table*}[t]
  \centering\scriptsize
  \setlength{\tabcolsep}{3pt}
  \begin{tabular}{@{}l*{4}{r@{\,}l}@{}}
\toprule
Estimate & \multicolumn{2}{c}{Qwen3-1.7B} & \multicolumn{2}{c}{OLMo} & \multicolumn{2}{c}{gpt-oss-20b} & \multicolumn{2}{c}{gpt-oss-120b} \\
\midrule
\multicolumn{9}{@{}l}{\textit{OLMo at its recommended decoding (Appendix~\ref{app:models})}} \\
Default's share above uniform$^{\dagger}$ & & & 42.4 & [39.4, 45.3] & & & & \\
MRV-like minus name-order share, held-out$^{\dagger}$ & & & 45.4 & [40.8, 50.4] & & & & \\
Breadth gain of randomised presentation, Symbolic$^{\dagger}$ & & & 45.3 & [42.1, 48.3] & & & & \\
\quad ZebraLogic$^{\dagger}$ & & & 21.1 & [15.1, 26.9] & & & & \\
\addlinespace[2pt]
\multicolumn{9}{@{}l}{\textit{Without the $B=2$ problems (Appendix~\ref{app:stats})}} \\
Validity change, random-choice instruction$^{a,\dagger}$ & & & $-$1.7 & [$-$3.2, $-$0.2] & & & & \\
\quad on the problems with $B=4$ or 6 & & & $-$0.8 & [$-$3.0, 1.6] & & & & \\
Validity change, temperature $+0.25$$^{a,\dagger}$ & & & $-$2.1 & [$-$3.6, $-$0.6] & & & & \\
\quad on the problems with $B=4$ or 6 & & & $-$1.3 & [$-$3.6, 1.1] & & & & \\
Olmo~3 7B Think, cued success: DPO minus SFT & \multicolumn{8}{c}{$-$0.4 [$-$1.0, 0.0]} \\
\quad on the problems with $B=4$ or 6 & \multicolumn{8}{c}{$-$0.8 [$-$1.6, $-$0.3]} \\
Olmo~3 7B Think, cued success: RLVR minus DPO & \multicolumn{8}{c}{0.3 [$-$0.1, 0.8]} \\
\quad on the problems with $B=4$ or 6 & \multicolumn{8}{c}{0.7 [0.0, 1.4]} \\
\addlinespace[2pt]
\multicolumn{9}{@{}l}{\textit{Search procedures (Appendix~\ref{app:procedures})}} \\
MRV-like minus name-order share, new classes & 38.0 & [30.4, 45.0] & 37.7 & [33.2, 42.3] & 25.3 & [19.3, 31.6] & 37.4 & [28.3, 46.5] \\
Candidates minus mentions, restated facts & 47.4 & [35.9, 58.3] & 55.7 & [46.6, 63.9] & 32.6 & [25.5, 39.5] & 44.6 & [34.3, 53.9] \\
\quad disjunctive fact & 3.9 & [$-$6.4, 13.9] & $-$9.2 & [$-$20.5, 2.3] & 16.5 & [7.1, 25.9] & 37.3 & [25.4, 48.9] \\
Static minus dynamic MRV share & 23.0 & [13.9, 31.9] & 8.5 & [$-$2.9, 18.8] & 1.5 & [$-$10.0, 11.6] & & \\
\addlinespace[2pt]
\multicolumn{9}{@{}l}{\textit{Scale and training stages (Appendix~\ref{app:stages})}} \\
Match to own procedure, minus gpt-oss-20b's & & & & & & & 9.1 & [6.4, 11.8] \\
Olmo~3 7B Think, DPO minus SFT: modal share & \multicolumn{8}{c}{5.9 [2.7, 9.2]} \\
\quad MRV-like contrast & \multicolumn{8}{c}{8.3 [2.2, 14.5]} \\
Olmo~3 7B Think, RLVR minus DPO: modal share & \multicolumn{8}{c}{$-$0.1 [$-$2.9, 2.7]} \\
\quad MRV-like contrast & \multicolumn{8}{c}{$-$2.8 [$-$8.6, 2.8]} \\
DPO minus SFT, prompt held fixed: modal share & \multicolumn{8}{c}{1.7 [$-$0.8, 4.2]} \\
\quad MRV-like contrast & \multicolumn{8}{c}{2.0 [$-$2.2, 6.3]} \\
DPO prompt minus SFT prompt: modal share & \multicolumn{8}{c}{4.22 [1.86, 6.64]} \\
\quad MRV-like contrast & \multicolumn{8}{c}{6.25 [1.78, 10.56]} \\
\addlinespace[2pt]
\multicolumn{9}{@{}l}{\textit{History (Appendix~\ref{app:history})}} \\
Greedy answer kept: its free share & & & 15.1 & [8.7, 21.2] & & & & \\
\quad cued success of the other solutions$^{\dagger}$ & & & 0.9 & [$-$0.1, 1.9] & & & & \\
Sampled alternative kept: its free share & & & 33.0 & [25.7, 40.3] & & & & \\
\quad cued success of the other solutions$^{\dagger}$ & & & $-$0.3 & [$-$1.2, 0.5] & & & & \\
\addlinespace[2pt]
\multicolumn{9}{@{}l}{\textit{Reduced reasoning$^{b}$ (Appendix~\ref{app:answer-only})}} \\
Default's share above uniform$^{\dagger}$ & 23.3 & [20.2, 26.4] & & & 23.8 & [20.9, 26.5] & 41.3 & [37.9, 44.7] \\
MRV-like minus name-order share, held-out$^{\dagger}$ & 8.7 & [3.6, 14.5] & & & 29.5 & [25.6, 33.5] & 44.0 & [38.0, 50.3] \\
\addlinespace[2pt]
\multicolumn{9}{@{}l}{\textit{Process evidence (Appendix~\ref{app:process})}} \\
Commitment rise at the first trial assignment & 44.3 & [32.0, 57.0] & 17.7 & [8.2, 28.3] & 11.2 & [0.5, 22.4] & & \\
Log reasoning length per dead end$^{c}$ & 0.013 & [$-$0.004, 0.027] & 0.023 & [0.007, 0.050] & 0.024 & [0.009, 0.042] & & \\
\quad in pairs differing in value labels$^{c}$ & & & 0.057 & [0.000, 0.107] & & & & \\
\addlinespace[2pt]
\multicolumn{9}{@{}l}{\textit{Breadth gain of randomised presentation minus that of a baseline (Appendix~\ref{app:baselines})}} \\
Verbalized Sampling, Symbolic & 23.4 & [18.6, 28.1] & 24.5 & [19.7, 29.2] & 24.1 & [19.6, 28.8] & & \\
Validity-preserving temperature, Symbolic & 26.9 & [22.6, 31.0] & 33.9 & [29.5, 37.9] & 27.9 & [23.6, 32.1] & & \\
Verbalized Sampling, ZebraLogic & 26.4 & [19.1, 33.5] & 17.4 & [6.0, 28.1] & $-$2.5 & [$-$7.9, 2.7] & $-$6.9 & [$-$14.1, 0.3] \\
Validity-preserving temperature, ZebraLogic & 23.9 & [18.3, 29.7] & 12.1 & [6.3, 18.0] & & & 17.5 & [11.9, 23.2] \\
\addlinespace[2pt]
\multicolumn{9}{@{}l}{\textit{Components of randomised presentation (Appendix~\ref{app:pa})}} \\
Breadth gain, value labels alone$^{\dagger}$ & 25.4 & [21.0, 30.0] & 36.2 & [31.6, 40.3] & 30.7 & [25.6, 35.4] & & \\
Breadth gain, fact order minus entity names$^{\dagger}$ & 1.9 & [$-$2.4, 6.3] & $-$8.3 & [$-$13.8, $-$3.0] & $-$9.2 & [$-$13.9, $-$4.0] & & \\
\addlinespace[2pt]
\multicolumn{9}{@{}l}{\textit{ZebraLogic after randomised presentation (Appendices~\ref{app:zebra-residual} and~\ref{app:zebra-source})}} \\
Default's share above the forecast & 11.9 & [6.7, 17.5] & 13.2 & [7.2, 19.1] & 11.2 & [6.7, 15.6] & & \\
Source solution's share above the forecast & 0.2 & [$-$5.0, 5.3] & 0.4 & [$-$5.5, 5.9] & 1.3 & [$-$3.7, 6.5] & & \\
\bottomrule
\end{tabular}
  \caption{\textbf{Secondary estimates with their intervals.}
  Estimates in points, each followed in brackets by its 95\% interval ($^{\dagger}$90\% interval, as stored with the analysis); a value set across the four model columns belongs to the Olmo~3 7B Think training stages.
  $^{a}$On 60 of the 72 problems, 40 of them with $B=4$ or 6 (Appendix~\ref{app:stats}).
  $^{b}$Qwen3-1.7B with thinking disabled; gpt-oss-20b and gpt-oss-120b at low reasoning effort.
  $^{c}$Change in log reasoning length, not in points.}
  \label{tab:secondary}
\end{table*}

\begin{table*}[t]
  \centering\scriptsize
  \setlength{\tabcolsep}{3pt}
  \begin{tabular}{@{}l*{6}{r@{\,}l}@{}}
\toprule
Condition and task & \multicolumn{2}{c}{Qwen3-1.7B} & \multicolumn{2}{c}{OLMo} & \multicolumn{2}{c}{gpt-oss-20b} & \multicolumn{2}{c}{gpt-oss-120b} & \multicolumn{2}{c}{Qwen3-30B-A3B} & \multicolumn{2}{c}{Gemma} \\
\midrule
\multicolumn{13}{@{}l}{\textit{Shuffled presentation}} \\
\quad Calendar & $-$9.0 & [$-$11.8, $-$6.4]$^{*}$ & $-$8.4 & [$-$11.6, $-$5.2]$^{*}$ & $-$8.0 & [$-$9.9, $-$6.0]$^{*}$ & $-$0.7 & [$-$1.3, $-$0.1] & $-$1.0 & [$-$2.0, $-$0.2] & $-$0.1 & [$-$0.4, 0.3] \\
\quad Colouring & $-$2.2 & [$-$5.8, 1.5]$^{v}$ & $-$4.1 & [$-$7.0, $-$1.1]$^{v}$ & $-$1.6 & [$-$2.9, $-$0.4] & $-$0.3 & [$-$0.8, 0.1] & 1.4 & [$-$0.7, 3.9] & $-$0.3 & [$-$0.5, 0.0] \\
\quad Trip & \multicolumn{2}{c}{--} & \multicolumn{2}{c}{--} & 1.5 & [$-$0.8, 4.3] & $-$0.8 & [$-$2.8, 1.2] & 0.3 & [$-$2.7, 3.7] & \multicolumn{2}{c}{--} \\
\addlinespace[2pt]
\multicolumn{13}{@{}l}{\textit{Random order}} \\
\quad Calendar: times & 15.9 & [10.5, 21.5] & $-$9.5 & [$-$13.2, $-$5.9]$^{v}$ & $-$4.6 & [$-$6.6, $-$2.4]$^{v}$ & $-$1.9 & [$-$2.7, $-$1.1] & $-$8.6 & [$-$11.7, $-$5.6]$^{v}$ & $-$6.3 & [$-$9.2, $-$3.7]$^{v}$ \\
\quad Trip: cities & \multicolumn{2}{c}{--} & \multicolumn{2}{c}{--} & 1.1 & [$-$0.3, 2.4] & $-$2.2 & [$-$4.0, $-$0.5] & $-$1.3 & [$-$3.2, 0.5] & \multicolumn{2}{c}{--} \\
\bottomrule
\end{tabular}
  \caption{\textbf{Validity changes behind the shuffled-presentation and random-order tests on the public planning tasks}: change in validity against the derived prompt, in points, with 90\% intervals.
  The test of Table~\ref{tab:benchmark-estimates} requires a breadth gain and a lower limit above $-5$ points here.
  $^{v}$Lower limit below $-5$ points while breadth rose, the cells marked $^{v}$ in Table~\ref{tab:benchmark-estimates}. $^{*}$Lower limit below $-5$ points, with no breadth gain either. --: not admitted or stopped at the pilot.}
  \label{tab:benchmark-validity}
\end{table*}

\paragraph{Figures~\ref{fig:interventions} and~\ref{fig:appendix-interventions}.}
Figure~\ref{fig:interventions} shows four interventions, and Figure~\ref{fig:appendix-interventions} repeats it with the two that were run on the Symbolic task with the main models only, randomised presentation with renamed entities and enumerate-then-choose.
The pre-specified test requires a breadth gain and a validity change whose 90\% lower limit lies above minus the margin, with Holm correction within each model's family of tests; the shading marks this one-sided region, and an open marker means that the test did not pass (Appendix~\ref{app:stats} gives the decisions under the two-sided tests originally specified).
Every comparison that has such a test is drawn with the estimates the test used: the breadth gain with its 95\% interval and the validity change with its 90\% interval.
For the two new models, randomised presentation and the random search order show the tested breadth gains and validity intervals.
The random-choice instruction, the higher temperature and enumerate-then-choose are descriptive comparisons without such a test.
On the Symbolic task they cover all 72 problems for the main models.
The two new models' random-choice comparison comes from a paired bootstrap over 51 clusters with 95\% intervals on both axes (Appendix~\ref{app:stats}); they were not run with renamed entities, the higher temperature or enumerate-then-choose, and gpt-oss-120b was run with randomised presentation only.
The validity axes are clipped at $-12.5$ points; in Figure~\ref{fig:appendix-interventions}, OLMo's validity with enumerate-then-choose is $-20.5$ points (90\% interval $[-22.7, -18.0]$) and is drawn as an arrow.
In panel B the earlier models' estimates for the random-choice instruction have no interval, and the higher-temperature comparison uses the original runs.

\section{Secondary estimates}
\label{app:secondary}

Table~\ref{tab:secondary} collects, with their intervals, the estimates of the secondary analyses that Appendices~\ref{app:models} to~\ref{app:zebra-original} report.
Those sections give each estimate and its decision in words, and print an interval only where a result is null or unresolved.
Intervals are at the level stored with each analysis: 95\%, or 90\% in the rows marked $^{\dagger}$.

Table~\ref{tab:benchmark-validity} gives the validity change of every shuffled-presentation and random-order test on the public planning tasks (Appendix~\ref{app:benchmarks}).

\end{document}